\documentclass[table]{formatting/gtech}

\usepackage[utf8]{inputenc}             
\usepackage{textcomp}                   
\def\BibTeX{{\rm B\kern-.05em{\sc i\kern-.025em b}\kern-.08em
    T\kern-.1667em\lower.7ex\hbox{E}\kern-.125emX}}

\tcbuselibrary{skins,breakable}
\usepackage{makecell}                  
\usepackage{array}                     
\usepackage{colortbl}                  
\usepackage{adjustbox}                 
\usepackage{arydshln}                  
\usepackage{algorithm}                 
\usepackage{algorithmicx}              
\usepackage[noend]{algpseudocode}      
\algrenewcommand\algorithmiccomment[1]{\hfill\textcolor{gray!80!black}{\scriptsize$\triangleright$ #1}} 

\usepackage{amssymb,amsfonts}
\usepackage{amsthm}
\theoremstyle{definition}

\usepackage[shortlabels]{enumitem}     
\setenumerate[1]{itemsep=2pt,partopsep=1pt,parsep=\parskip,topsep=3pt}
\setitemize[1]{itemsep=2pt,partopsep=1pt,parsep=\parskip,topsep=3pt}
\setdescription{itemsep=2pt,partopsep=1pt,parsep=\parskip,topsep=3pt}

\usepackage{balance}                   
\usepackage{siunitx}                   
\usepackage[normalem]{ulem}            
\useunder{\uline}{\ul}{}               
\usepackage{pifont}                    
\usepackage{tikz}                      

\newtcolorbox{finding}{
    colback=gray!8,                    
    colframe=black!40,                 
    boxrule=0.4pt,                     
    arc=1.5pt,                         
    left=2pt, right=2pt, top=2pt, bottom=2pt,
    before skip=5pt, after skip=5pt,
}

\newcolumntype{C}[1]{>{\centering\arraybackslash}m{#1}}
\newcolumntype{R}[1]{>{\raggedleft\arraybackslash}m{#1}}
\newcolumntype{L}[1]{>{\raggedright\arraybackslash}p{#1}}
\newcolumntype{I}{!{\vrule}}
\newcommand{\LATTEArena}{\textsf{LATTEArena}}
\newcommand{\CAAFE}{\textsf{CAAFE}}
\newcommand{\SMARTFEAT}{\textsf{SMARTFEAT}}
\newcommand{\FeatLLM}{\textsf{FeatLLM}}
\newcommand{\OCTree}{\textsf{OCTree}}
\newcommand{\FEBias}{\textsf{FEBias}}
\newcommand{\GPTS}{\textsf{GPT-Signal}}
\newcommand{\FEBP}{\textsf{FEBP}}
\newcommand{\ELLM}{\textsf{ELLM-FT}}
\newcommand{\FREEFORM}{\textsf{FREEFORM}}
\newcommand{\LLMFE}{\textsf{LLM-FE}}
\newcommand{\Adda}{\textsf{Adda}}
\newcommand{\LFG}{\textsf{LFG}}
\newcommand{\LPFG}{\textsf{LPFG}}
\newcommand{\RAFG}{\textsf{RAFG}}
\newcommand{\Rouge}{\textsf{Rouge\ One}}
\newcommand{\GRFG}{\textsf{GRFG}}
\newcommand{\Autofeat}{\textsf{Autofeat}}

\newcommand{\CoTNL}{\texttt{CGN}}
\newcommand{\CoTCode}{\texttt{CGC}}
\newcommand{\CoTCodeh}{$\texttt{CGC}_{\texttt{h}}$}
\newcommand{\CoTCodet}{$\texttt{CGC}_{\texttt{t}}$}
\newcommand{\CoTCodec}{$\texttt{CGC}_{\texttt{c}}$}
\newcommand{\CoTNLh}{$\texttt{CGN}_{\texttt{h}}$}
\newcommand{\CoTNLt}{$\texttt{CGN}_{\texttt{t}}$}
\newcommand{\CoTRPN}{\texttt{CGR}}
\newcommand{\CoTRPNh}{$\texttt{CGR}_{\texttt{h}}$}
\newcommand{\CoTRPNt}{$\texttt{CGR}_{\texttt{t}}$}

\newcommand{\ToTNL}{\texttt{TMN}}
\newcommand{\ToTCode}{\texttt{TMC}}
\newcommand{\ToTNLh}{$\texttt{TMN}_{\texttt{h}}$}
\newcommand{\ToTCodeh}{$\texttt{TMC}_{\texttt{h}}$}
\newcommand{\ToTRPN}{\texttt{TMR}}
\newcommand{\ToTRPNh}{$\texttt{TMR}_{\texttt{h}}$}
\newcommand{\CriticRPN}{\texttt{GGR}}
\newcommand{\CriticCode}{$\texttt{GGC}$}
\newcommand{\CriticNL}{$\texttt{GGN}$}
\newcommand{\OPRO}{$\texttt{OGC}$}
\newcommand{\OPROC}{$\texttt{OGC}_{\texttt{c}}$}
\newcommand{\OPRORPN}{$\texttt{OGR}$}
\newcommand{\OPRONL}{$\texttt{OGN}$}
\newcommand{\EvoRPNw}{$\texttt{EBR}_{\texttt{w}}$}
\newcommand{\EvoRPNr}{$\texttt{EBR}_{\texttt{r}}$}

\newcommand{\EvoRPN}{\texttt{EBR}}
\newcommand{\EvoCode}{\texttt{EBC}}

\newcommand{\circled}[1]{\tikz[baseline=(char.base)]{
    \node[shape=circle,draw,inner sep=0pt] (char) {#1};}}

\newcommand{\mynote}{\textcolor{blue!70!black}{$\blacklozenge$}} 

\newtcolorbox{summarybox}{
  colback=gray!5!white,     
  colframe=gray!30,         
  boxrule=0.5pt,            
  arc=3pt,                  
  left=5pt, right=5pt, top=3pt, bottom=3pt, 
  before skip=5pt,          
  after skip=5pt
}

\definecolor{highlightcol}{RGB}{232, 245, 253}  
\definecolor{highlightcol2}{RGB}{210, 233, 250}
\definecolor{NLColor}{RGB}{232, 245, 253}    
\definecolor{RPNColor}{RGB}{255, 245, 200}   
\definecolor{CodeColor}{RGB}{210, 235, 210}  
\definecolor{blue1}{RGB}{65,120,190}   
\definecolor{blue2}{RGB}{95,150,210} 
\definecolor{blue3}{RGB}{140,180,230}
\definecolor{yellow1}{RGB}{210,180,80}
\definecolor{yellow2}{RGB}{230,210,130}
\definecolor{yellow3}{RGB}{250,240,180}



\newtheoremstyle{elegantdef}
  {1.5ex plus 0.2ex minus 0.2ex} 
  {1.5ex plus 0.2ex minus 0.2ex} 
  {\itshape}                   
  {}                           
  {\bfseries\color{blue1!85!black}} 
  {.}                          
  {.5em}                       
  {}                           

\theoremstyle{elegantdef}
\newtheorem{definition}{Definition}[section]

\crefname{definition}{Definition}{Definitions}
\Crefname{definition}{Definition}{Definitions}

\title{\LATTEArena{}: An Evaluation Framework for LLM-powered Tabular Feature Engineering (Extended Version)}
\author[1]{Ankai Hao}
\author[1]{Ke Chen}
\author[1]{Huan Li}
\author[1]{Lidan Shou}
\affiliation[1]{Zhejiang University}

\gtechdata[Keywords]{LLM for feature engineering, benchmarking}

\abstract{
Feature engineering remains a cornerstone of tabular data analysis, and Large Language Models (LLMs) have emerged as a promising paradigm for its automation, giving rise to \emph{LLM-powered Automated Tabular Feature Engineering} (LATTE). However, the field lacks standardized, cost-aware evaluation platforms, and the combinatorial explosion of design choices obscures true algorithmic progress. To bridge these gaps, we systematically deconstruct 15 representative LATTE methods into a unified 6-dimensional taxonomy. Based on this abstraction, we introduce \LATTEArena{}, a standardized, modular, and extensible benchmarking framework that decouples monolithic pipelines into reusable execution blocks. By distilling the massive combinatorial space, we evaluate 24 core LATTE configurations across 7 research questions. Our head-to-head benchmarking goes beyond predictive accuracy to quantify token efficiency and execution robustness, yielding 17 empirical findings on cost-effectiveness trade-offs. Furthermore, we provide 3 concrete recommendations for optimal real-world deployment. By enabling controlled component-level comparisons, \LATTEArena{} shifts the paradigm from ad-hoc prompt engineering to systematic context management. All code, datasets, and over 4,000 execution logs are publicly available to foster a dynamic, community-driven benchmark. Our framework, leaderboard, and all artifacts are hosted on the LATTEArena project website at \url{https://goodenhak.github.io/LATTEArena/}.
}

\begin{document}

\begingroup
\centering
\fontsize{8}{10}\selectfont\color{gray}
\noindent This work has been submitted to the IEEE for possible publication.
Copyright may be transferred without notice, after which this version may no longer be accessible.\par
\endgroup
\vspace{0.4em}

\maketitle

\section{Introduction}\label{sec:introduction}

The ``Garbage In, Garbage Out'' principle dictates that \emph{data quality} fundamentally bounds AI model performance. Feature engineering~\cite{10.5555/3239815}, the critical bridge between raw data and algorithms, remains indispensable for tabular data despite deep learning's success in reducing manual feature design for computer vision and natural language processing tasks. Tabular data dominates high-stakes domains such as recommendation systems, healthcare, and finance, where tree-based models consistently outperform deep learning in efficiency and interpretability, further amplifying the need for high-quality features. With approximately 80\% of data science effort devoted to data cleaning and feature engineering~\cite{shen2018automated}, automating this labor-intensive pipeline is paramount, cementing Tabular Automated Feature Engineering (TAFE) as a focal point in the AutoML era.

Historically, TAFE relied on heuristic search~\cite{8215494,7344858,khurana2016cognito}, meta-learning~\cite{10.5555/3172077.3172240}, and reinforcement learning~\cite{khurana2018feature,chen2019neural,zhu2022difer}, which often suffer from high computational costs and limited ability to discover semantically complex features. The advent of Large Language Models (LLMs), with their robust semantic understanding and code synthesis capabilities, has transformed this landscape~\cite{wang2026toward}. A pivotal breakthrough was \CAAFE{}~\cite{hollmann2023large}, which leveraged Chain-of-Thought~\cite{wei2022chain} prompting with \texttt{GPT-4}~\cite{achiam2023gpt} to iteratively generate features. This success catalyzed a surge of \textbf{L}LM-powered \textbf{A}u\textbf{T}omated \textbf{T}abular feature \textbf{E}ngineering (\textbf{LATTE}) methods, encompassing diverse prompt-based~\cite{lin2023smartfeat,ijcai2025p782,nam2024optimized}, search~\cite{ijcai2025p782}, and RAG-enhanced~\cite{Zhang2024RetrievalAugmentedFG} paradigms, evolving even into fully autonomous data science workflows~\cite{li2024autokaggle}.

\begin{figure}
  \centering
  \includegraphics[width=0.8\linewidth]{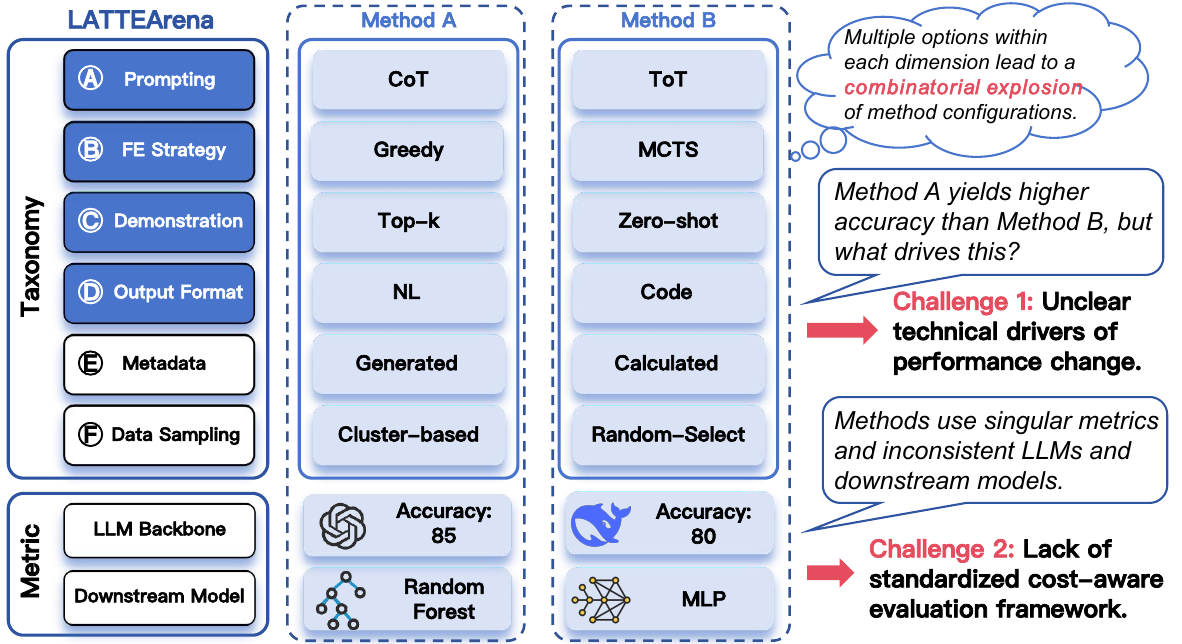}
  \caption{The \LATTEArena{} taxonomy and challenges.}
  \label{fig:taxonomy_challenges}
\end{figure}

Despite this rapid emergence, LATTE significantly lags behind other LLM-driven tabular tasks like table question answering and NL2SQL. As depicted in \Cref{fig:taxonomy_challenges}, our survey of 15 representative LATTE methods reveals that the community confronts two critical bottlenecks. \textbf{\emph{Challenge 1: Combinatorial explosion obscures performance attribution.}} Existing methods are proposed as monolithic pipelines, arbitrarily bundling heterogeneous choices from prompting paradigms, search strategies, output formats, and other dimensions. This structural entanglement makes it impossible to isolate which specific components actually drive performance gains versus which merely add overhead. A systematic, component-level decomposition across all dimensions is therefore essential to determine optimal combinations.
\textbf{\emph{Challenge 2: Absence of standardized, cost-aware evaluation.}} Existing literature relies on fragmented and inconsistent experimental setups, varying arbitrarily across datasets, downstream models, and base LLMs. This discrepancy prohibits fair, head-to-head comparisons. Furthermore, evaluations exclusively focus on predictive accuracy, neglecting real-world deployment metrics such as token consumption, inference latency, and method robustness. Consequently, the critical trade-offs between performance gains and practical costs remain obscured.

To break through these bottlenecks, there is a compelling need to systematically deconstruct these monolithic methods and evaluate their underlying components head-to-head. To this end, we introduce \textbf{\LATTEArena{}}, the first standardized, modular, and execution-safe benchmarking platform for LLM-powered feature engineering. By abstracting the LATTE design space into modular execution blocks, \LATTEArena{} enables seamless integration and controlled ablation of any novel technique, effectively shifting the research paradigm from ad-hoc prompt tuning to systematic context engineering~\cite{mei2025survey}. Specifically, our contributions span three progressive layers, from theoretical abstraction through engineering infrastructure to empirical insights:

\begin{itemize}[leftmargin=*]
\item \textbf{Unified Taxonomy (\Cref{sec:taxonomy})}: 
We systematically deconstruct \textbf{15} recent, representative LATTE methods into a comprehensive \textbf{6}-dimensional taxonomy, theoretically abstracting the sprawling combinatorial design space into modular, comparable components. Based on this abstraction, we compile representative approaches and curate the \LATTEArena{} datasets (\Cref{ssec:data}).

\item \textbf{Standardized Evaluation Framework (\Cref{sec:bench})}: 
We instantiate \LATTEArena{} as an extensible execution infrastructure. By rigorously distilling the massive design space based on cost-effectiveness principles, we standardly implement and head-to-head benchmark \textbf{24} core configurations that effectively proxy the entire LATTE landscape.

\item \textbf{Actionable Empirical Insights (\Cref{sec:exp,sec:recommendation})}: 
We conduct massive controlled benchmarking across \textbf{7} core research questions. Going beyond accuracy to quantify token efficiency and execution robustness, we extract empirical findings on optimal design choices and provide concrete recommendations for real-world deployment. All code, datasets, and over \textbf{4,000} replayable execution logs are publicly released to foster a community-driven arena.
\end{itemize}

\section{Preliminaries}
\label{sec:preliminaries}

\subsection{Concepts and Definitions}

\begin{definition}[\textbf{Tabular Dataset}]\label{definition:tabular_dataset}
A \emph{tabular dataset} is denoted as $\mathcal{D}^{(M+1)\times N}$, containing $(M+1)$ columns (comprising $M$ feature columns and one label column) and $N$ rows (instances). 
The dataset is partitioned row-wise into disjoint subsets for training, validation, and testing: $
\mathcal{D} = \{\mathcal{D}_{\text{train}},\, \mathcal{D}_{\text{valid}},\, \mathcal{D}_{\text{test}}\}.
$
Since the tabular data semantics are invariant to column permutation, the label column is conventionally placed at the end of the feature set without loss of generality.
Let the feature matrix be $\mathbf{X} \in \mathbb{R}^{N\times M}$ and the label vector be $\mathbf{y} \in \mathbb{R}^N$. 
Each row $\mathbf{x}_i \in \mathbb{R}^{1 \times M}$ corresponds to a single instance, and each column $\mathbf{f}_j \in \mathbb{R}^{N \times 1}$ corresponds to one feature across all instances. 
The dataset can thus be equivalently expressed in row-wise and column-wise forms:
\begin{equation}\label{equ:dataset}
\mathcal{D}^{(M+1)\times N} 
= \{\mathbf{X},\, \mathbf{y}\} 
= \{(\mathbf{x}_i,\, y_i)\}_{i=1}^N 
= \{\mathbf{f}_1,\, \ldots,\, \mathbf{f}_M,\, \mathbf{y}\}.
\end{equation}
\end{definition}

\begin{definition}[Supervised Learning on Tabular Data]\label{definition:supervised_learning_tabular_dataset}
Given a training tabular dataset $\mathcal{D}_{\text{train}} = \{(\mathbf{x}_i, y_i)\}_{i=1}^{N'}$, a predictive model built from supervised learning is defined as:
\begin{equation}\label{equ:prediction_model}
\hat{y}_i \gets \mathcal{H}(\mathbf{x}_i; \theta),
\end{equation}
where $\mathcal{H}(\cdot;\theta)$ denotes a function parameterized by learnable parameters $\theta$, mapping input features to a prediction $\hat{y}_i$.
The optimal parameters are obtained by minimizing the empirical risk:
\begin{equation}\label{equ:loss_function}
\theta^{*} \gets \arg\min\nolimits_{\theta} \mathcal{L}(\theta, \mathcal{D}) =  \arg\min\nolimits_{\theta} \frac{1}{{N'}} \sum\nolimits_{i=1}^{N'} 
\ell\!\left(\mathcal{H}(\mathbf{x}_i; \theta), y_i\right),
\end{equation}
where $\ell(\cdot)$ is the loss function, commonly cross-entropy for classification~\cite{hollmann2025accurate,bonet2024hyperfast} or mean squared error for regression~\cite{yan2024making}.
\end{definition}

\begin{definition}[Automated Feature Engineering on Tabular Data]\label{definition:autoFE}
Typically, a tabular supervised learning pipeline places a feature engineering module before a downstream predictive model. 
Such a feature engineering process is defined as a parameterized transformation over the dataset $\mathcal{D} = \{\mathcal{D}_{\text{train}}, \mathcal{D}_{\text{valid}}, \mathcal{D}_{\text{test}}\}$:
\begin{equation}
\mathcal{D}_{\phi} \gets \{\phi(\mathbf{X}),\, \mathbf{y}\}, \quad \phi \in \Phi,
\end{equation}
where $\phi$ represents a sequence of feature engineering operations within a search space $\Phi$ (e.g., feature selection, transformation, or embedding).

The objective of \emph{automated feature engineering} (AutoFE) is to identify the transformation $\phi^{*}$ that yields the best downstream performance, formulated as a bilevel optimization problem:
\begin{equation}\label{equ:optimization}
\begin{aligned}
\min\nolimits_{\phi \in \Phi} \; & \mathcal{L}\!\left(\theta^{*}(\phi),\, \mathcal{D}_{\text{valid},\phi}\right) \\
\text{s.t.} \quad & 
\theta^{*}(\phi) \in \arg\min\nolimits_{\theta \in \Theta} \, 
\mathcal{L}\!\left(\theta, \, \mathcal{D}_{\text{train},\phi} \right).
\end{aligned}
\end{equation}
The inner optimization learns the model parameters $\theta^{*}(\phi)$ on the feature-transformed training data $\mathcal{D}_{\text{train},\phi}$, while the outer optimization selects the transformation $\phi^{*}$ that minimizes the validation loss. 
Evaluation is then performed on $\mathcal{D}_{\text{test},\phi^{*}}$ using the pair $\{ \phi^{*},\, \theta^{*}(\phi^{*}) \}$.
\end{definition}

\subsection{The LATTE Pipeline}
\label{section2.2}

Recent advances in LLMs have opened new opportunities for AutoFE on tabular data. 
Unlike traditional AutoFE methods that focus primarily on statistical analysis, LLM-based approaches~\cite{hollmann2023large,lin2023smartfeat,wang-etal-2024-gpt} leverage the semantic understanding of features.  
By integrating contextual and semantic knowledge, \textbf{\emph{LLM-powered Automated Tabular Feature Engineering}} (LATTE) enables more intelligent transformation design, helping to generate meaningful features and reduce redundancy.
The formal definition of LATTE is provided as follows.

\begin{definition}[LLM-powered AuTomated Tabular feature Engineering, LATTE]
\label{definition:latte}
LATTE aims to automatically generate and refine feature transformations using an LLM as the optimization engine.
At each iteration $i = 1,\ldots,S$, its feature optimization module $\mathcal{P}_{\mathcal{M}}$, implemented by an LLM $\mathcal{M}$, takes as input the training data $\mathcal{D}_{\text{train}}$, validation data $\mathcal{D}_{\text{valid}}$, 
previous context $\mathcal{C}_{i-1}$, and historical feature engineering operations $\Phi_{i-1}$, and outputs a new operation sequence $\phi_i$ with context $\mathcal{C}_i$: 
\begin{equation}
\{ \phi_i, \mathcal{C}_i \} \gets \mathcal{P}_{\mathcal{M}}(\mathcal{D}_{\text{train}},\, \mathcal{D}_{\text{valid}},\, \mathcal{C}_{i-1},\, \Phi_{i-1}),
\quad i = 1,\ldots,S,
\end{equation}
where $\Phi_{i-1} = \{\phi_1, \phi_2, \ldots, \phi_{i-1}\}$ accumulates all generated operations.

LATTE follows the same bilevel structure as the automated feature engineering described in~\Cref{equ:optimization}.
In essence, the LLM iteratively proposes feature transformations guided by validation feedback, progressively improving downstream model performance.
\end{definition}

\begin{figure}
  \centering
  \includegraphics[width=0.8\linewidth]{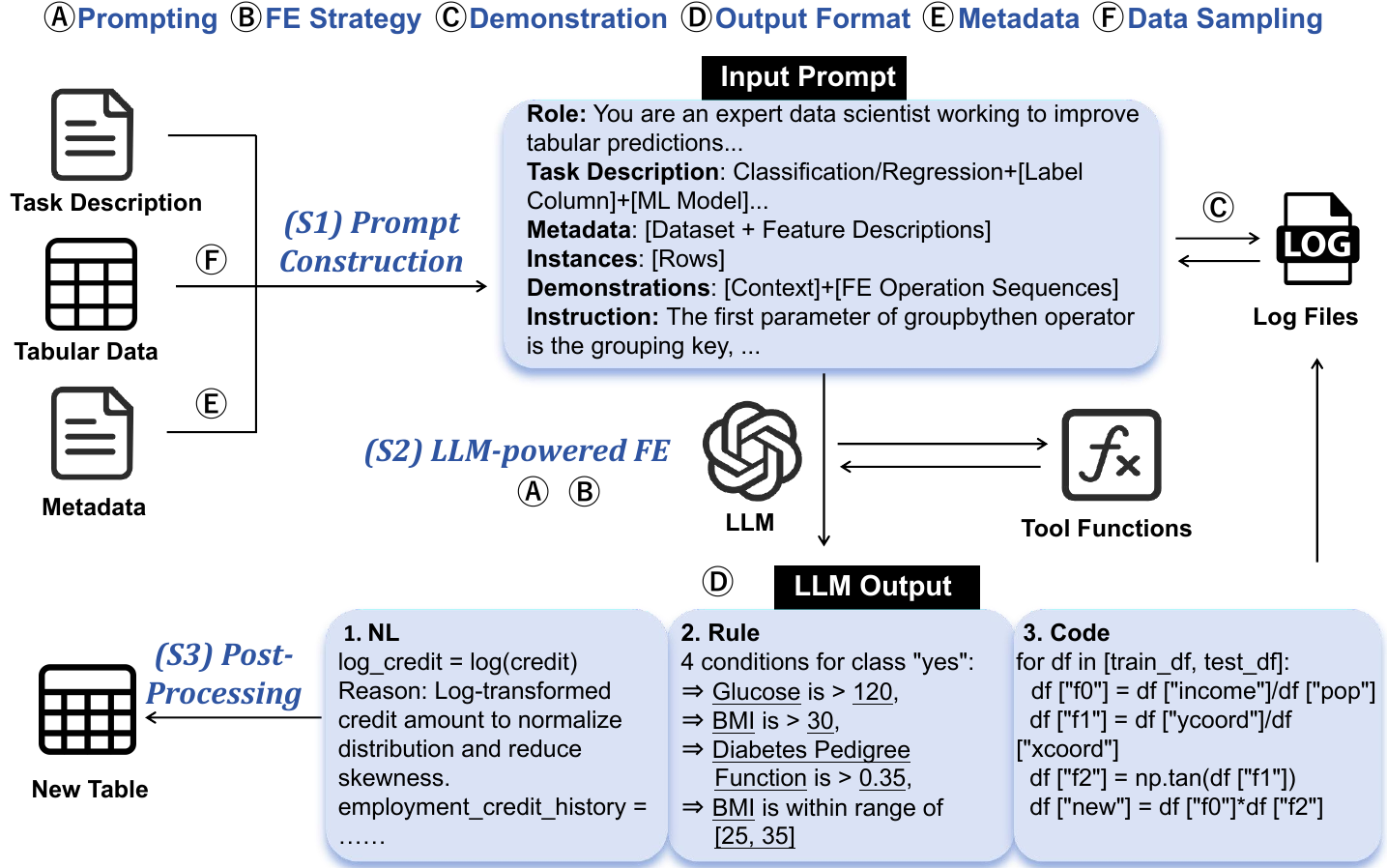}
  \caption{The LATTE pipeline and 6-dimensional taxonomy. Based on our analysis, dimensions \protect\circled{A}-\protect\circled{B}, \protect\circled{C}-\protect\circled{D}, and \protect\circled{E}-\protect\circled{F} have major, medium, and minor impacts on performance, respectively.}
  \label{fig:pipeline}
\end{figure}

As shown in~\Cref{fig:pipeline}, the LATTE pipeline consists of three main stages:
\begin{enumerate}[leftmargin=*, label=\textbf{(S\arabic*)}]
\item \textbf{Prompt Construction}. 
The LATTE pipeline begins with the design of high-quality prompts that guide the LLM toward effective feature generation. 
Referring to the example in~\Cref{fig:pipeline}, each prompt typically comprises six key components: 
(i) \emph{role}, which defines the LLM's perspective;
(ii) \emph{task description}, which specifies the prediction type (e.g., classification or regression), label column name, and downstream model to be optimized;
(iii) \emph{metadata}, which summarizes dataset and feature information, optionally including distributional statistics pre-computed from the tabular data;
(iv) \emph{instances}, which provide representative table rows for contextual grounding;
(v) \emph{demonstrations}, which are retrieved from log files and contain context–operation sequence pairs $(\mathcal{C}, \phi)$, where $\mathcal{C}$ stores condensed historical metadata and reasoning traces to reduce storage overhead;
and (vi) \emph{instructions}, which typically define the expected output format and offer guidance on how feature engineering operators should be applied.

\item \textbf{LLM-powered Feature Engineering}.
This stage forms the core of the LATTE pipeline. 
The feature optimizer $\mathcal{P}_{\mathcal{M}}$, composed of the LLM backbone $\mathcal{M}$ and supporting tool functions, determines candidate transformations through iterative querying.
For each dataset version, the algorithm selects appropriate prompts and updates based on its feature engineering strategy, which may follow greedy, evolutionary, or UCB-based search paradigms. 
At this stage, all modifications affect only the stored context $\mathcal{C}$ and operation sequence $\phi$, each of which corresponds to a specific dataset version.  
Advanced prompting techniques such as Chain-of-Thought (CoT), self-consistency, and Retrieval-Augmented Generation (RAG)~\cite{gao2024retrievalaugmentedgenerationlargelanguage} can be incorporated to enhance reasoning quality and diversity in the generated transformations.

\item \textbf{Post-processing}.
Since the LLM outputs feature transformations in textual form, the post-processing module converts them into executable operations.  
The output $\phi$ may appear in natural language (NL), rule, Reverse Polish notation (RPN), or code format, guided by the input instructions and demonstrations.  
The parser translates $\phi$ into executable programs, applies the transformations to produce a new dataset and metadata, and records the resulting $(\mathcal{C}, \phi)$ pairs in log files for further iterations.
\end{enumerate}

\section{Taxonomy}\label{sec:taxonomy}

Despite their apparent diversity, existing LATTE methods share a compact set of orthogonal design axes.
Building on the three-stage pipeline (\Cref{fig:pipeline}), we propose a \textbf{6-dimensional taxonomy} synthesizing 15 representative LATTE methods published since \CAAFE{}~\cite{hollmann2023large}. As presented in \Cref{tab:commands}, we prioritize these dimensions (\mbox{\circled{A}}-\mbox{\circled{F}}) based on their technical diversity and relative impact on performance. For each dimension, we analyze trade-offs of its technical options to establish a rigorous rationale for our benchmarking choices.

\begin{table*}[t]
  \caption{Taxonomy of 15 recent representative LATTE methods; categorized by the core Prompting paradigm (CoT, ToT, OPRO, etc.). Dark blue/light blue/none shading denotes major/medium/minor performance impacts. \textcolor{gray}{Gray options} are excluded from benchmarking due to scalability or controllability limits; ``$\smallsetminus$'' means absent.}
  \label{tab:commands}
  \footnotesize
  \centering
  \setlength{\aboverulesep}{0.6ex}
  \setlength{\belowrulesep}{0.6ex}
  \setlength{\tabcolsep}{2pt}
  \renewcommand{\arraystretch}{1}
  \resizebox{\textwidth}{!}{
  \begin{tabular}{
    c
    c
    c
    >{\columncolor{highlightcol2}}c
    >{\columncolor{highlightcol}}C{3.3cm}
    c
    C{2.0cm}
    >{\columncolor{highlightcol2}}c
    >{\columncolor{highlightcol}}c}
    \toprule
    \multirow{2}{*}{\textbf{Family}} &
    \multirow{2}{*}{\textbf{Methods}} &
    \multirow{2}{*}{\textbf{Venue}} & 
    \multicolumn{4}{c}{\textbf{Prompt Construction}} &
    \multicolumn{1}{c}{\textbf{LLM-powered FE}} & 
    \multicolumn{1}{c}{\textbf{Post-processing}} \\
    \cmidrule(lr){4-7}\cmidrule(lr){8-8}\cmidrule(lr){9-9}
    & & & 
    \multicolumn{1}{>{\columncolor{highlightcol2}}c}{\circled{A} Prompting} & 
    \multicolumn{1}{>{\columncolor{highlightcol}}c}{\circled{C} Demonstration} & 
    \multicolumn{1}{c}{\circled{E} Metadata} & 
    \multicolumn{1}{c}{\circled{F} Data Sampling} & 
    \multicolumn{1}{>{\columncolor{highlightcol2}}c}{\circled{B} FE Strategy} & 
    \multicolumn{1}{>{\columncolor{highlightcol}}c}{\circled{D} Output Format}\\
    \midrule
    \multirow{7}{*}{\rotatebox{0}{\textbf{CoT}}} 
    & \CAAFE{}~\cite{hollmann2023large} & NeurIPS 2024 & Vanilla CoT  & \textcolor{gray}{Full Context} & \textcolor{gray}{Human-Written} & \textcolor{gray}{Random-Selected} & Greedy Incremental & Code \\
    & \FEBias{}~\cite{kuken2024large}& NeurIPSW 2024 & Vanilla CoT & \textcolor{gray}{Full Context} & Calculated Value & \textcolor{gray}{Random-Selected} & Greedy Incremental & NL\\
    & \GPTS{}~\cite{wang-etal-2024-gpt}&FINNLP 2024 & Vanilla CoT & $\smallsetminus$ & LLM-Generated & \textcolor{gray}{Human-Selected} & Greedy Incremental & NL \\
    & \RAFG{}~\cite{Zhang2024RetrievalAugmentedFG}&ICDM 2025 & Vanilla CoT & $\smallsetminus$ & \textcolor{gray}{RAG-Enhanced} & $\smallsetminus$ & Greedy Incremental & Code\\
    & \SMARTFEAT{}~\cite{lin2023smartfeat}& CIDR 2024 & \textcolor{gray}{Operator-based CoT} & $\smallsetminus$ & Native & $\smallsetminus$ & Greedy Incremental & NL \\
    & \FeatLLM{}~\cite{han2024large}& ICML 2024 & CoT + \textcolor{gray}{SC} & $\smallsetminus$ & Native & Cluster-based & \textcolor{gray}{Select-Expand-Ensemble} & Rule \\
    & \FREEFORM{}~\cite{lee2025knowledge}&AMIA 2025 & CoT + \textcolor{gray}{SC} & \textcolor{gray}{Human-Written NL Features} & $\smallsetminus$ & \textcolor{gray}{Random-Selected} & \textcolor{gray}{Select-Expand-Ensemble} & NL \\
    \midrule
    \multirow{2}{*}{\rotatebox{0}{\textbf{ToT}}} 
    & \LFG{}~\cite{ijcai2025p782}& IJCAI 2025 & ToT & Positive-Negative Features & Native & $\smallsetminus$ & MCTS-based Search & NL \\
    & \Adda{}~\cite{lu2025adda}& SIGMOD 2025 & ToT & Top-$k$ Code Snippets & Calculated Value & \textcolor{gray}{Random-Selected} & MCTS-based Search & Code \\
    \midrule
    \multirow{2}{*}{\rotatebox{0}{\textbf{OPRO}}} 
    & \OCTree{}~\cite{nam2024optimized}&NeurIPS 2024  & CART-based OPRO & Top-$k$ (Code, CART, Score) & Native & $\smallsetminus$ & Greedy Incremental & Code \\
    & \FEBP{}~\cite{zou2025automated}&Preprint 2025 & Vanilla OPRO & Top-$k$ (RPN, Score) & Native & $\smallsetminus$& \textcolor{gray}{Expand-Reduce} & RPN \\
    \midrule
    \multirow{2}{*}{\rotatebox{0}{\textbf{Evo}}} 
    & \ELLM{}~\cite{gong2025evolutionary}& AAAI 2025 & EvoPrompt & Ranked (RPNSet, Score) & $\smallsetminus$ & $\smallsetminus$& Best-of-N & RPN \\
    & \LLMFE{}~\cite{abhyankar2025llm}& Preprint 2025  & EvoPrompt & Top-$k$ Code Snippets & Native & \textcolor{gray}{Random-Selected} & Best-of-N & Code\\
    \midrule
    \multirow{2}{*}{\rotatebox{0}{\textbf{Critic}}} 
    & \LPFG{}~\cite{ijcai2025p314}& IJCAI 2025 & Generator-Critic & Textual Gradient & Calculated Value & $\smallsetminus$ & Greedy Incremental & RPN \\
    & \Rouge{}~\cite{bradland2025knowledge} & Preprint 2025 & Generator-Critic & Textual Gradient & \textcolor{gray}{RAG-Enhanced} & $\smallsetminus$ & Greedy Incremental & Code\\
    \bottomrule
  \end{tabular}}
\end{table*}

\subsection[Prompting Techniques (Major)]{Prompting Techniques \mbox{\protect\circled{A}} (Major)}
\label{ssec:prompting}

In LATTE, prompting techniques function less as reasoning templates and more as \textbf{\emph{search policies}} over the feature space.
Their key differentiator is the trade-off among exploration breadth, feedback granularity, and query overhead.
We identify five families that form a spectrum from single-trajectory refinement to closed-loop multi-agent collaboration.

\subsubsection{Chain of Thought (CoT)}
CoT~\cite{brown2020language,10.5555/3600270.3602070} decomposes complex problems into intermediate reasoning steps, serving as the \emph{baseline} from which all other LATTE prompting techniques generalize.
\textbf{\emph{Vanilla CoT}}, the simplest variant, guides the LLM through analytical steps before deriving feature operations.
\textbf{\emph{Operator-based CoT}}~\cite{lin2023smartfeat} decomposes the problem further by querying the LLM twice per round (first for operator selection, then for operand specification), thereby constraining the search space at the cost of additional queries.
\textbf{\emph{CoT with Self-Consistency (SC)}} generates $k$ independent reasoning paths and aggregates outcomes. Notably, LATTE adapts SC differently from standard NLP practice~\cite{wang2022self}: rather than voting on operation sequences $\{\phi_1^*, \ldots, \phi_k^*\}$, it produces $k$ transformed datasets, trains $k$ models, and ensembles predictions~\cite{han2024large,lee2025knowledge}. This shifts cost from exploration to ensembling, improving robustness at significantly higher training overhead.

\subsubsection{Tree of Thought (ToT)}
\textbf{\emph{ToT}}~\cite{yao2023tree} generalizes CoT from a single trajectory to a branching tree of reasoning paths, enabling broader exploration of the feature space.
\LFG{}~\cite{ijcai2025p782} instantiates ToT for LATTE by assigning each tree node a candidate $\phi$ evaluated via validation performance.
A critical distinction from ToT in logic or math tasks is the absence of a natural termination condition: LATTE exploration must be bounded by depth or step limits rather than by a verifiable solution, making the exploration budget a first-order hyperparameter.

\subsubsection{EvoPrompt} 
\textbf{\emph{EvoPrompt}}~\cite{guo2023connecting} casts prompt optimization as evolutionary search, treating prompts as individuals and performance as fitness.
In LATTE, however, only the dynamic segment (few-shot demonstrations) is evolved while the static template remains fixed, making EvoPrompt effectively a \emph{demonstration-quality optimizer} rather than a full prompt optimizer.
\ELLM{}~\cite{gong2025evolutionary} and \LLMFE{}~\cite{abhyankar2025llm} further depart from classical evolutionary algorithms by replacing \texttt{mutation} and \texttt{crossover} with unconstrained LLM rewrites, making the process highly dependent on $\mathcal{M}$'s intrinsic capabilities. Since each population requires a separate LLM query per round, this incurs substantial overhead that scales with population size.

\subsubsection{Optimization by PROmpting (OPRO)}
OPRO~\cite{yang2023large} frames the LLM as a black-box optimizer, iteratively refining outputs using an objective function as feedback.
\textbf{\emph{Vanilla OPRO}}~\cite{zou2025automated} uses validation loss directly, generating candidate features each round and retaining the best.
\textbf{\emph{CART-based OPRO}}~\cite{nam2024optimized} replaces raw metadata with reasoning extracted from a trained Classification and Regression Tree, explicitly surfacing important columns and prediction thresholds. This bridges statistical and semantic metadata, offering the LLM interpretable feedback that goes beyond scalar loss values.

\subsubsection{Generator-Critic}
This multi-agent paradigm separates generation and evaluation into two interacting agents that form a closed feedback loop.
In \LPFG{}~\cite{ijcai2025p314}, the critic diagnoses the current feature set and conditions the generator via \emph{textual gradients}, which provide semantic and distributional advice functioning as a differentiable signal in text space.
\Rouge{}~\cite{bradland2025knowledge} distributes the critic role across multiple specialized agents, whose evaluations collectively steer subsequent generations.
Among all five families, Generator-Critic provides the tightest feedback coupling, but at the cost of requiring at least two LLM calls per iteration.

\begin{summarybox}
\textbf{Remarks:} The five prompting families form a spectrum from single-trajectory refinement ({CoT}) through branching exploration ({ToT}), population-based search ({EvoPrompt}), and feedback-driven optimization ({OPRO}), to closed-loop multi-agent collaboration ({Generator-Critic}). Moving along this spectrum generally increases exploration capability and feedback quality, but with \emph{diminishing returns} relative to escalating query costs, an empirical finding we quantify in~\Cref{ssec:TEA}. We benchmark \textbf{all five paradigms}: \textcolor{gray}{Operator-based CoT} is normalized into {CoT} as it mainly changes prompt syntax, while \textcolor{gray}{CoT+SC} is not isolated as its effect is entangled with downstream ensembling.
\end{summarybox}

\subsection[Feature Engineering Strategies (Major)]{Feature Engineering Strategies \mbox{\protect\circled{B}} (Major)}
\label{ssec:feature_engineering}

\begin{figure}
  \centering
  \includegraphics[width=\linewidth]{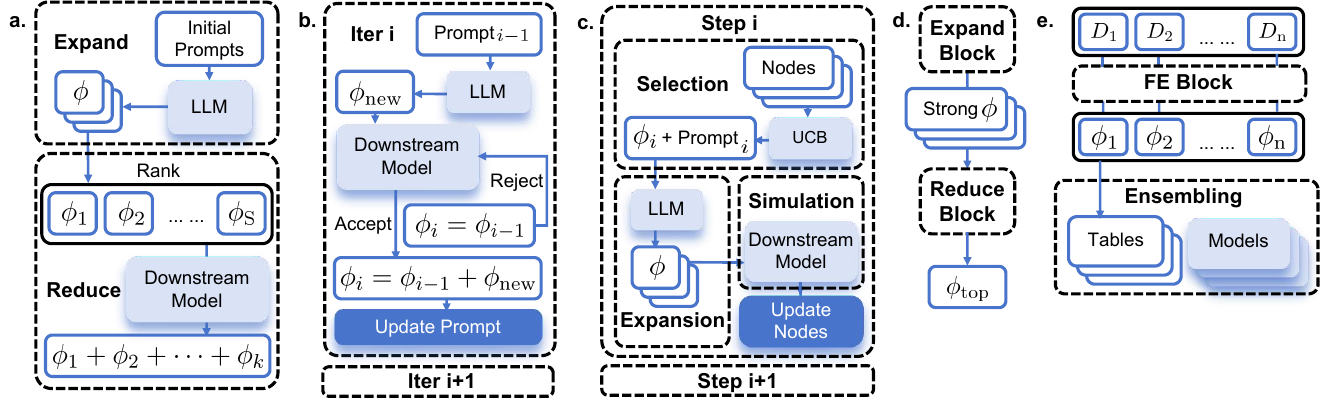}
  \caption{The LATTE feature engineering strategies: a. Expand-Reduce, b. Greedy Incremental, c. MCTS-based Search, d. Best-of-N, e. Select-Expand-Ensemble.}
  \label{fig:Strategy}
\end{figure}

While prompting techniques determine \emph{how} $\mathcal{M}$ is queried, the FE strategy determines \emph{what to do} with the resulting $\phi$: whether to adopt it, how to compose it with prior operations, and when to stop searching.
We identify five strategies, each instantiating a different point in the exploration--exploitation trade-off (\Cref{fig:Strategy}). 
For clarity, we denote the validation loss as $ \text{loss}(\phi) = \mathcal{L}\!\left(\theta^*(\phi), \mathcal{D}_{\text{val},\phi}\right) $, and $\phi_1+\phi_2=\operatorname{Concat}(\phi_1,\phi_2)$.

\subsubsection{Expand-Reduce (\Cref{fig:Strategy}a)}
This two-phase strategy, inherited from classical AutoFE~\cite{8215494,7344858,7837936}, fully decouples generation from selection.
The \emph{Expansion} phase queries $\mathcal{M}$ $S$ times to produce a candidate pool, with no feedback between queries.
The \emph{Reduction} phase then applies feature selection (e.g., Forward Selection~\cite{zou2025automated}) to identify the best subset $\phi^*$.
This modularity is both its strength (each phase can be optimized independently) and its weakness: the generation phase receives no performance signal, potentially producing many low-quality candidates. Moreover, exact subset selection requires $2^S$ model trainings; greedy approximations reduce this to $O(S)$ but ignore feature interactions.

\subsubsection{Greedy Incremental (\Cref{fig:Strategy}b)}
The most widely adopted strategy (used by 8 of 15 methods), Greedy Incremental operates as a hill-climbing algorithm over the feature space.
At each iteration $i$, $\mathcal{P}_{\mathcal{M}}$ generates $\phi_\text{new}$ and adopts it only if $\text{loss}(\phi_{i-1}) > \text{loss}(\phi_{i-1}+\phi_\text{new})$, thereby incorporating inter-feature interactions into every decision.
Its dominance reflects practical simplicity: one query and one model evaluation per round.
However, the greedy acceptance criterion makes it vulnerable to local optima, a fundamental limitation that motivates the tree-based strategies below.

\subsubsection{MCTS-based Search (\Cref{fig:Strategy}c)}
This strategy combines ToT with Monte Carlo Tree Search~\cite{coulom2006efficient} and the Upper Confidence Bound~\cite{kocsis2006bandit} (UCB) to principally balance exploration and exploitation.
Each tree node maintains an operation sequence $\phi$ with its UCB score:
\begin{equation}
    \text{UCB}(i) = \frac{1}{n} \sum_{j=1}^{n} \bigl(\text{loss}(\phi_i) - \text{loss}(\phi_{\text{d}_j})\bigr) + \alpha \sqrt{\frac{2 \ln(v_{\text{p}})}{v_i}},
\end{equation}
where $d_j$ indexes descendant nodes, $v_i$ and $v_{\text{p}}$ are visit counts, and $\alpha$ controls the exploration--exploitation balance.
Unlike standard MCTS, the simulation step always appends $\phi_\text{new}$ regardless of loss change, resembling progressive widening rather than rollout-based evaluation.
This provides principled exploration beyond greedy methods but at substantially higher cost: each expansion requires $k$ queries for a $k$-ary tree, plus full UCB backpropagation.

\subsubsection{Best-of-N (\Cref{fig:Strategy}d)}
A simplified Expand--Reduce variant that forgoes the reduction phase entirely, selecting $\phi_\text{top}=\arg\min \text{loss}(\phi)$ from $N$ candidates.
This strategy places the entire burden on generation quality, making it viable only when paired with prompt-optimization techniques (typically EvoPrompt~\cite{abhyankar2025llm,gong2025evolutionary}) that raise the baseline quality of individual generations.
Its practical advantage is low downstream overhead: the simplified selection avoids the combinatorial cost of subset search.

\subsubsection{Select-Expand-Ensemble (\Cref{fig:Strategy}e)}
The only strategy that explores the \emph{data space} rather than the feature space alone.
It first partitions $\mathcal{D}$ into subsets by label or feature relevance, independently generates $\phi$ for each, and merges the results.
Some methods additionally apply a reduction phase; e.g., \FREEFORM{}~\cite{lee2025knowledge} uses LLM scoring for feature selection.
Existing implementations defer merging to the prediction stage, ensembling downstream models trained on each subset~\cite{lee2025knowledge,han2024large}. This introduces data-space diversity orthogonal to the feature-space exploration of other strategies.

\begin{summarybox}
\textbf{Remarks:} {Greedy Incremental} dominates current practice, reflecting its simplicity, despite its known vulnerability to local optima. We evaluate \textbf{Greedy Incremental}, \textbf{MCTS} (for principled exploration), and \textbf{Best-of-N} (for candidate diversification). We exclude \textcolor{gray}{Expand-Reduce} due to prohibitive combinatorial selection costs, and \textcolor{gray}{Select-Expand-Ensemble} as its gains stem from downstream ensembling rather than inherent feature search. The field currently lacks strategies that combine efficient exploration with low cost.
\end{summarybox}

\subsection[Demonstration Composition (Medium)]{Demonstration Composition \mbox{\protect\circled{C}} (Medium)}\label{ssec:demos}
Demonstrations are the primary mechanism for inter-iteration knowledge transfer.
The candidate pool for round $k$ consists of historical records $\{(\phi_i, C_i)\}_{i=1}^{k-1}$, each typically comprising an operation sequence $\phi$, a score derived from $\text{loss}(\phi)$, and LLM-generated reasoning in $\mathcal{C}$.
The core design question is: \emph{what should the LLM remember from its history?}
We identify five strategies that form a spectrum from raw replay to increasingly abstract summarization.

\textbf{\emph{Ranked / Top-$k$}}. Candidates are ranked by score and the top $k$ are retained, controlling prompt length while biasing toward high-quality $\phi$~\cite{gong2025evolutionary,abhyankar2025llm,zou2025automated}.
In \Adda{}~\cite{lu2025adda}, ranking uses a composite metric combining semantic similarity (cosine distance between metadata embeddings) and structural similarity (tree edit distance~\cite{10.5555/338219.338628} on feature lineage), enabling cross-dataset transfer at the cost of offline embedding training.

\textbf{\emph{Positive-Negative}}. Candidates are partitioned into successful (loss-decreasing) and unsuccessful groups~\cite{ijcai2025p782}, providing the LLM with contrastive signal about what works and what does not.

\textbf{\emph{Full Context}}. Early methods such as \CAAFE{} leverage multi-turn dialogue to grant LLMs access to the complete history. While maximizing information, this approach scales poorly: context length grows linearly with iteration count, eventually exceeding window limits.

\textbf{\emph{Textual Gradient}}. The most abstract approach: a dedicated critic agent distills history into high-level guidance (e.g., \emph{semantic advice}, \emph{distributional advice}~\cite{ijcai2025p314}, or \emph{focus areas}~\cite{bradland2025knowledge}) which is injected into the generator's prompt. This creates an abstraction barrier between memory and generation, conceptually analogous to gradient descent in text space.

\begin{summarybox}
\textbf{Remarks:} From Full Context through Top-$k$ and Positive-Negative to Textual Gradient, there is a clear trend toward more compressed, higher-level history representations. Among these strategies, \textcolor{gray}{Full Context} is excluded as its cost grows sharply with iteration history; \textcolor{gray}{Human-Written}, while effective, is excluded as it introduces non-scalable expert priors and hinders fair comparison. The key challenge is distilling stable guidance from locally valid observations, a problem closely related to meta-learning, yet no current method explicitly addresses it.
\end{summarybox}

\subsection[Output Format (Medium)]{Output Format \mbox{\protect\circled{D}} (Medium)}
\label{ssec:format}
The output format determines both the \emph{reachable feature space} and the dominant \emph{failure mode}, making it a first-order design choice. 
Prior work shows that increasing formal and syntactic constraints often degrades LLM reasoning quality~\cite{tam-etal-2024-speak}, creating a fundamental tension between expressiveness and reliability.
We identify four formats along this trade-off axis.

\textbf{\emph{Natural Language (NL)}}. Free-form text descriptions of feature operations. Maximally expressive and transparent, but the lack of structural constraints makes parsing fragile and limits NL methods to low-order operations per round~\cite{lin2023smartfeat,lee2025knowledge,kuken2024large}.

\textbf{\emph{Rule}}. Decision-tree-style rules where each rule maps to a binary feature~\cite{han2024large,nam2024optimized}. Offers structural clarity but is inherently limited to discrete features.

\textbf{\emph{Code}}. Python programs that directly implement FE operations~\cite{hollmann2023large,abhyankar2025llm,lu2025adda}. This format is highly expressive and executable, but introduces strong syntactic and environmental constraints, making runtime failures the dominant risk.

\textbf{\emph{Reverse Polish Notation (RPN)}}. Postfix operation sequences (e.g., $f_1; f_2; +; \text{square}$) that enable unambiguous, deterministic execution~\cite{zou2025automated,gong2025evolutionary,ijcai2025p314}. RPN avoids Code's runtime fragility while supporting high-order operations within a single round, expanding the explorable feature space. The trend is notable: three 2025 methods adopt RPN versus none in 2023--2024, suggesting convergence toward formal but compact representations.

\begin{summarybox}
\textbf{Remarks:} \textbf{NL} is limited in expressing complex features, whereas \textbf{Code} is expressive but error-prone. \textbf{RPN} balances compositional expressiveness with execution reliability. The recent shift toward RPN reflects a pragmatic preference for reliability over maximal expressiveness. We quantify this format--performance--robustness trade-off in~\Cref{ssec:PGA,ssec:robust}. \textcolor{gray}{Rule} is excluded as it is tightly coupled to tree-style binary predicates and cannot represent many general continuous or high-order transformations.
\end{summarybox}

\subsection[Metadata Construction (Minor)]{Metadata Construction \mbox{\protect\circled{E}} (Minor)}
\label{ssec:meta}
Metadata, which encompasses descriptive information about table structure, provenance, and feature semantics, is the primary channel through which domain knowledge enters the LATTE pipeline, distinguishing it from statistically-driven AutoFE.
However, existing methods vary widely in how much semantic grounding they provide.
We identify five categories based on how metadata is obtained or enriched, ordered from static to dynamic.

\textbf{\emph{Native}}. Metadata residing inherently within source databases or extracted directly from web repositories. Used by a majority of methods (6/15), it is the lowest-cost option but is limited to whatever schema information the data source provides.

\textbf{\emph{Human-Written}}. Precise attribute descriptions and data-type specifications manually curated by data engineers~\cite{hollmann2023large}. High-fidelity but non-scalable.

\textbf{\emph{LLM-Generated}}. When native metadata is insufficient, LLMs are queried to generate or augment descriptions from feature names~\cite{wang-etal-2024-gpt}. Quality depends on the LLM's domain coverage.

\textbf{\emph{Calculated Value}}. Statistical measures (e.g., mean, kurtosis, mutual information) computed from the data itself. Libraries such as PyMFE~\cite{JMLR:v21:19-348} automate this across General, Statistical, and Information-Theoretic categories. These values complement semantic metadata by exposing data complexity and distribution patterns invisible from column names alone.

\textbf{\emph{RAG-Enhanced}}. Metadata enriched via Retrieval-Augmented Generation, enabling domain-specific grounding~\cite{Zhang2024RetrievalAugmentedFG,bradland2025knowledge}. This is the most expressive option but introduces retrieval latency and corpus-quality dependencies.

\begin{summarybox}
\textbf{Remarks:} Metadata is the main lever for semantic grounding, yet remains underexplored: 6/15 methods rely on \textbf{Native} metadata. Our experiments (\Cref{ssec:PMA}) show that metadata quality significantly affects feature generation quality, suggesting that dynamic enrichment (i.e., \textbf{Calculated Value}, \textbf{LLM-Generated}) is a promising but underexploited direction. \textcolor{gray}{Human-Written} metadata is non-scalable, while \textcolor{gray}{RAG-Enhanced} metadata introduces retrieval-quality and latency confounders.
\end{summarybox}

\subsection[Data Sampling Method (Minor)]{Data Sampling Method \mbox{\protect\circled{F}} (Minor)}\label{ssec:sample}
Because tabular datasets often exceed LLM context limits, sampling determines which rows the model actually sees, making it a \emph{context-compression} problem rather than a miniature version of full-data analysis.
The goal is to expose representative feature-label patterns under strict token budgets.
Existing methods adopt one of three approaches, each encoding a different prior about what constitutes an informative sample.

\textbf{\emph{Random-Selected}}. The simplest approach: a uniformly sampled instance set. Makes no assumptions about data structure but wastes context budget on potentially redundant rows.

\textbf{\emph{Cluster-based}}. Introduced by \FeatLLM{}~\cite{han2024large}, this approach clusters instances by label and selects representatives from each cluster, ensuring the LLM observes class-specific patterns. Effective when label boundaries carry the most structural information.

\textbf{\emph{Human-Selected}}. Domain experts manually identify representative instances~\cite{wang-etal-2024-gpt}, typically in high-cardinality or heavily imbalanced domains (e.g., finance, biology) where automated clustering is impractical.

\begin{summarybox}
\textbf{Remarks:} All strategies are either label-driven or agnostic; no method optimizes sampling for \emph{feature informativeness} (e.g., selecting rows that maximize feature variance or minimize redundancy). Given tight context budgets, feature-aware sampling represents an open direction. We use \textbf{Cluster-based} for controlled benchmarking: \textcolor{gray}{Random-Selected} may waste context on redundant rows, while \textcolor{gray}{Human-Selected} introduces expert priors and weak reproducibility.
\end{summarybox}

\section{\LATTEArena{}: Design and Usage}\label{sec:bench}

\begin{figure*}
    \centering
    \includegraphics[width=1.0\linewidth]{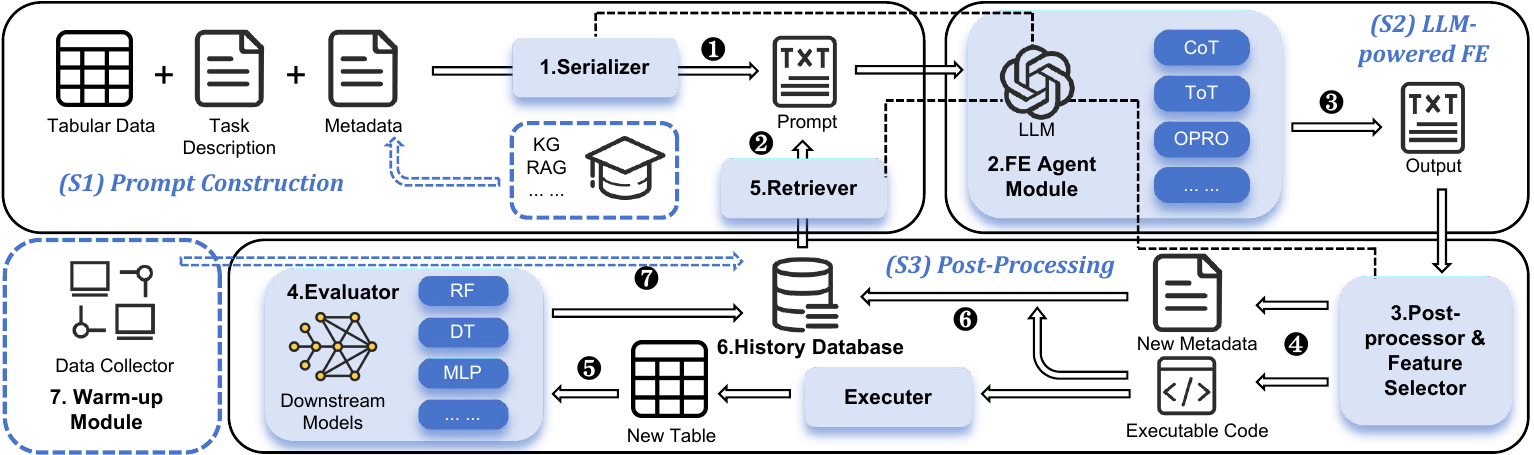}
    \caption{\LATTEArena{} pipeline. The architecture bridges the conceptual taxonomy in~\Cref{sec:taxonomy} with an execution-safe implementation. Blue dashed boxes indicate optional modules (RAG, Warm-up) that can be adaptively routed based on configuration.}
    \label{fig:benchmark}
\end{figure*}

\subsection{Motivation and Abstract Framework}
While~\Cref{sec:taxonomy} conceptualizes the LATTE design space, empirical validation remains fundamentally hindered by the monolithic nature of existing methods. Current implementations tightly couple their core algorithmic innovations (e.g., prompting strategies) with fragile, dataset-specific execution environments. This obscures performance attribution, making it impossible to isolate whether empirical gains stem from a search strategy or merely from over-engineered heuristics.

To establish a scientifically rigorous evaluation methodology, we introduce \LATTEArena{}, the first standardized, modular, and execution-safe framework designed for LATTE, as illustrated in~\Cref{fig:benchmark}. The framework organizes the pipeline into three stages as shown in~\Cref{fig:pipeline}: Prompt Construction, LLM-powered FE, and Post-processing. These stages are realized through seven core modules: Serializer, FE Agent, Post-processor and Feature Selector, Evaluator, Retriever, History Database, and Warm-up Module. All modules conform to standardized input-output specifications, enabling interchangeable implementation of the techniques surveyed in~\Cref{sec:taxonomy}. This modular architecture offers three principal benefits. First, \textbf{\emph{High-level Abstraction}}: the pipeline abstracts away syntactic idiosyncrasies of diverse prompting strategies, output formats, and LLM backends, effectively decoupling algorithmic search from execution logic. Second, \textbf{\emph{Seamless Extensibility}}: new techniques and modules from the six-dimensional taxonomy can be plugged in or adaptively routed without modifying the underlying backbone. Third, \textbf{\emph{Execution Safety}}: \LATTEArena{} sanitizes and robustifies LLM outputs, reducing the runtime failures that plague code-generation methods.

The pipeline adopts an iterative workflow. In Stage 1) \emph{\uline{Prompt Construction}}, the Serializer ingests task specifications, metadata, and tabular data to instantiate prompt templates based on technique configuration (\ding{202}). It supports diverse prompting strategies through a unified prompt template library covering techniques in~\Cref{ssec:prompting}, operators used in existing works, and output formats in~\Cref{ssec:format}. The Retriever queries the History Database for prior FE records and assembles demonstrations using strategies in~\Cref{ssec:demos} (\ding{203}). 

In Stage 2) \uline{\emph{LLM-powered FE}}, the FE Agent processes assembled prompts, where the optimizer $\mathcal{P}_\mathcal{M}$ selects prompts and dataset versions from candidates conditioned on the current input and FE strategy described in~\Cref{ssec:feature_engineering}. $\mathcal{P}_\mathcal{M}$ then uses the prompting techniques in~\Cref{ssec:prompting} to query $\mathcal{M}$, obtaining text segments containing operation sequences $\phi$ and context $\mathcal{C}$ (\ding{204}).

In Stage 3) \uline{\emph{Post-processing}}, the Post-processor converts the FE Agent outputs into executable code and updated metadata (\ding{205}). Subsequently, the Feature Selector filters all features using criteria such as mutual information and feature importance, and correspondingly updates the code and metadata. While prior studies primarily focus on prompting strategies, they often overlook output format compliance, which is essential for automatic parsing and execution. We define the \textbf{\emph{Success Rate}} as the fraction of LLM outputs that can be correctly parsed and executed without runtime errors. \LATTEArena{} improves this metric by enforcing stricter format constraints during prompt construction and enabling LLM-based reformatting, yielding a more robust and automation-oriented pipeline.

The Evaluator then scores features using downstream models referred to as \textbf{\emph{validation models}} (\ding{206}). As discussed in~\Cref{sec:taxonomy}, most methods use validation loss as the score. Furthermore, \LATTEArena{} integrates NAS~\cite{luo2018neural} and HPO~\cite{yu2020hyper} to enable realistic evaluations across diverse downstream models within an AutoML setting.

The History Database archives metadata, code, and evaluation scores from each iteration (\ding{207}). Together with the Retriever, it implements a preliminary form of context management~\cite{mei2025survey}, which is largely overlooked by existing LATTE methods that primarily focus on improving the FE Agent. 

Despite its advantages, \LATTEArena{} suffers from a \emph{cold start} problem: the initially empty History Database leaves early prompts without demonstrations. The Warm-up Module mitigates this by pre-populating the database using RL-based TAFE algorithms~\cite{li2023learning,10.5555/3618408.3620168} (\ding{208}), enabling few-shot learning from the first iteration and accelerating convergence of the iterative pipeline.

\subsection{Revisiting Existing Methods}\label{ssec:revisit}
Based on our taxonomy (\Cref{sec:taxonomy}), any existing LATTE approach can be decomposed into a specific configuration along six orthogonal dimensions (\Cref{tab:commands}). This decomposition is necessary because current LATTE methods differ not only in their advertised algorithmic ideas, but also in many auxiliary implementation choices, including output parsers, feature selectors, metadata serializers, demonstrations, and evaluators. These hidden differences make direct reproduction-based comparison misleading: a method may appear stronger because it uses a more robust selector or richer metadata rather than because its prompting or search policy is intrinsically better. Moreover, several representative methods~\cite{han2024large,ijcai2025p782} do not incorporate feature selection mechanisms, placing them at an inherent disadvantage in evaluations.
With the proposed taxonomy and the \LATTEArena{} framework, different technical combinations can be implemented via simple pipeline configuration, enabling controlled and fair comparisons across methods. The goal is therefore not to duplicate every original codebase verbatim, but to standardize them under common execution, selection, and evaluation interfaces while preserving their core algorithmic choices. However, naively evaluating every possible combination exposes a massive, intractable \textbf{\emph{combinatorial design space}}. For instance, multiplying just the primary options in~\Cref{tab:commands} (5 prompting strategies, 5 FE strategies, 5 demonstration choices, and 4 output formats) yields 500 unique pipelines.

\subsubsection{The Combinatorial Explosion Problem}
To systematically distill this intractable space into a meaningful benchmark, we introduce three \textbf{\emph{Configuration Principles}}. These principles identify meaningful and representative configurations by systematically comparing, consolidating, and selecting among existing techniques. The quantitative experimental results validating these principles are detailed in~\Cref{ssec:CLA}:
\begin{enumerate}[leftmargin=*, label=\textbf{(\arabic*)}]
\item \uline{\emph{Focusing on Core Paradigms.}}
Selected techniques should represent methodologically distinct approaches rather than stylistic variations (e.g., minor rephrasing of prompts). We further exclude techniques dependent on external resources (e.g., human-in-the-loop, RAG/KG, pre-trained embedding models) to assess the intrinsic capabilities of standalone LATTE methods faithfully. Model ensembling strategies (e.g., Select-Expand-Ensemble) are evaluated separately under the AutoML setting rather than included in the main configuration space.

\item \uline{\emph{Respecting Component Constraints.}}
Technique combinations must respect inherent design constraints and employ validated defaults. For instance, MCTS is paired with ToT, as MCTS was proposed as an improvement over Greedy Search specifically for ToT; EvoPrompt requires Best-of-N for multi-population evolution; OPRO and EvoPrompt are paired with output formats that can represent higher-order features; and Warm-up is limited to RPN due to constraints imposed by the RL data collector.

\item \uline{\emph{Prioritizing Cost-Effectiveness.}}
While advanced techniques like Least-to-Most prompting~\cite{zhouleast} excel in general reasoning tasks, their complexity introduces nontrivial token and computational overhead in the LATTE domain. We therefore prioritize a systematic exploration of simpler configurations, as they may reveal more cost-effective solutions that deliver comparable performance.
\end{enumerate}

\subsubsection{Resulting Configuration Space}\label{sssec:config}
Applying these principles reduces the design space to \textbf{24 core configurations}. They can be mapped to existing LATTE methods, serving as proxies on the \LATTEArena{}, as shown in~\Cref{tab:revisiting}. These configurations span four primary dimensions: prompting technique, FE strategy, demonstration, and output format, ensuring comprehensive coverage of key design choices. The remaining two dimensions adopt fixed defaults: metadata is set to Calculated Value and data sampling to Cluster-based, with metadata variations examined separately in~\Cref{ssec:PMA}. These configurations integrate and improve upon implementations of original methods, thus satisfying Principle 2. The CoT and ToT Families have richer configurations enabled by Principle 3.

\begin{table*}[htbp]
\small
\centering
\caption{\LATTEArena{} configuration space and mapping to original methods. Aliases are formed by concatenating the initials of their corresponding dimensions. For example, \texttt{CGN} represents a configuration combining \texttt{C}oT, \texttt{G}reedy search, and \texttt{N}L output. The Positive-Negative history demonstration uses the subscript `\texttt{h}'.}
\label{tab:revisiting}
\renewcommand{\arraystretch}{1.9}
\setlength{\tabcolsep}{3pt}
\resizebox{1.0\textwidth}{!}{
\begin{tabular}{lIc|c|c|cIcIl}
\toprule
\multirow{2}{*}{\textbf{Alias}} & \multicolumn{4}{cI}{\textbf{\LATTEArena{} Configuration}} & \multirow{2}{*}{\textbf{Original Methods}} & \multirow{2}{*}{\textbf{Detailed Differences of Each Method}} \\
\cmidrule(lr){2-5} 
& \textbf{Prompting} & \textbf{Strategy} & \textbf{Demonstration} & \textbf{Output} & & \\
\midrule
\CoTNL{}    & \texttt{C}oT & \texttt{G}reedy & $\smallsetminus$ & \texttt{N}L & \multirow{3}{*}{\makecell[l]{\SMARTFEAT{}~\cite{lin2023smartfeat}, \\ \GPTS{}~\cite{wang-etal-2024-gpt}, \\ \RAFG{}~\cite{Zhang2024RetrievalAugmentedFG}, \\ \FeatLLM{}~\cite{han2024large},\\\FREEFORM{}~\cite{lee2025knowledge}}} & \multirow{3}{*}{\makecell[l]{
\SMARTFEAT{} is a user-interactive dialogue system that uses 3 feature selection\\
metrics provided by sklearn. \GPTS{} has no open-source implementation\\
and is also a semi-automatic method involving human participation. \FREEFORM{}\\
and \FeatLLM{} are model ensemble methods directly oriented toward prediction\\
tasks, where the former targets linear classifiers and the latter is designed for\\
genotype data; \RAFG{} leverages RAG for assistance.}}\\
\CoTCode{}  & \texttt{C}oT & \texttt{G}reedy & $\smallsetminus$ & \texttt{C}ode & & \\
\CoTRPN{}   & \texttt{C}oT & \texttt{G}reedy & $\smallsetminus$ & \texttt{R}PN & & \\
\cdashline{1-7}

\CoTNLh{}   & \texttt{C}oT & \texttt{G}reedy & Positive-Negative (\texttt{h}) & \texttt{N}L & \multirow{3}{*}{\makecell[l]{\FEBias{}~\cite{kuken2024large}, \\ \CAAFE{}~\cite{hollmann2023large}}} & \multirow{3}{*}{\makecell[l]{\FEBias{} has no open-source implementation and no selector.\\\CAAFE{} relies solely on LLMs for feature selection and lacks a selector.}}\\
\CoTCodeh{} & \texttt{C}oT & \texttt{G}reedy & Positive-Negative (\texttt{h}) & \texttt{C}ode & & \\
\CoTRPNh{}  & \texttt{C}oT & \texttt{G}reedy & Positive-Negative (\texttt{h}) & \texttt{R}PN & & \\
\cdashline{1-7}

\CoTNLt{}   & \texttt{C}oT & \texttt{G}reedy & \texttt{t}op-$k$ & \texttt{N}L & \multirow{3}{*}{\textcolor{blue1}{New Variant}} & \\
\CoTCodet{} & \texttt{C}oT & \texttt{G}reedy & \texttt{t}op-$k$ & \texttt{C}ode & & \\
\CoTRPNt{}  & \texttt{C}oT & \texttt{G}reedy & \texttt{t}op-$k$ & \texttt{R}PN & & \\

\midrule

\ToTNLh{}   & \texttt{T}oT & \texttt{M}CTS & Positive-Negative (\texttt{h}) & \texttt{N}L & \multirow{3}{*}{\makecell[l]{\LFG{}~\cite{ijcai2025p782}, \\ \Adda{}~\cite{lu2025adda}}} & \multirow{3}{*}{\makecell[l]{\LFG{} lacks a selector and invokes the LLM Agent through multi-turn conversations,\\
which leads to context length overflow issues when dealing with a large number of\\
features or rich metadata. \Adda{} requires pre-training a metadata embedding model\\
on datasets in advance, and leverages UDFs to integrate the LATTE algorithm into\\
the DBMS for acceleration.}}\\
\ToTCodeh{} & \texttt{T}oT & \texttt{M}CTS & Positive-Negative (\texttt{h}) & \texttt{C}ode & & \\
\ToTRPNh{}  & \texttt{T}oT & \texttt{M}CTS & Positive-Negative (\texttt{h}) & \texttt{R}PN & &  \\
\cdashline{1-7}

\ToTNL{}    & \texttt{T}oT & \texttt{M}CTS & $\smallsetminus$ & \texttt{N}L & \multirow{3}{*}{\textcolor{blue1}{New Variant}} & \\
\ToTCode{}  & \texttt{T}oT & \texttt{M}CTS & $\smallsetminus$ & \texttt{C}ode & & \\
\ToTRPN{}   & \texttt{T}oT & \texttt{M}CTS & $\smallsetminus$ & \texttt{R}PN & & \\

\midrule

\CriticNL{}  & \texttt{G}enerator-critic & \texttt{G}reedy & Positive-Negative (\texttt{h}) & \texttt{N}L & \multirow{3}{*}{\makecell[l]{\LPFG{}~\cite{ijcai2025p314}, \\ \Rouge{}~\cite{bradland2025knowledge}}} & \multirow{3}{*}{\makecell[l]{\Rouge{} does not have an open-source implementation and introduces external\\
knowledge through RAG, thus it is represented by \LPFG{}.}}\\
\CriticCode{}& \texttt{G}enerator-critic & \texttt{G}reedy & Positive-Negative (\texttt{h}) & \texttt{C}ode & & \\
\CriticRPN{} & \texttt{G}enerator-critic & \texttt{G}reedy & Positive-Negative (\texttt{h}) & \texttt{R}PN & & \\

\midrule

\OPROC{}    & CART-based \texttt{O}PRO & \texttt{G}reedy & \texttt{t}op-$k$ + \texttt{C}ART & \texttt{C}ode & \OCTree{}~\cite{nam2024optimized} & \multirow{3}{*}{\makecell[l]{No modification to \OCTree{}.\\ \FEBP{} does not have an open-source implementation and lacks detailed descriptions\\
for its implementation. To ensure a fair comparison, we implemented it by modifying\\
the \OCTree{} framework.}} \\
\cdashline{1-6}
\OPRO{}     & \texttt{O}PRO & \texttt{G}reedy & \texttt{t}op-$k$ & \texttt{C}ode & \textcolor{blue1}{New Variant} & \\
\cdashline{1-6}
\OPRORPN{}  & \texttt{O}PRO & \texttt{G}reedy & \texttt{t}op-$k$ & \texttt{R}PN & \FEBP{}~\cite{zou2025automated} & \\

\midrule

\EvoRPNw{}  & \texttt{E}voPrompt & \texttt{B}est-of-N & Ranked + \texttt{w}arm-up & \texttt{R}PN & \ELLM{}~\cite{gong2025evolutionary} & \multirow{3}{*}{\makecell[l]{\ELLM{}, as a continuation of the RL method \GRFG{}, only receives numerical tables\\
as input without incorporating metadata and instances, and it also does not do feature\\
selection. \LLMFE{} does not actually perform evolutionary algorithms but always\\
selects top-k demonstrations; \LATTEArena{} uses the population evolution framework\\
of \ELLM{} as a replacement.}}\\
\cdashline{1-6}
\EvoRPN{}   & \texttt{E}voPrompt & \texttt{B}est-of-N & Ranked & \texttt{R}PN & \textcolor{blue1}{New Variant} & \\
\cdashline{1-6}
\EvoCode{}  & \texttt{E}voPrompt & \texttt{B}est-of-N & Ranked & \texttt{C}ode & \LLMFE{}~\cite{abhyankar2025llm} & \\

\bottomrule
\end{tabular}}
\end{table*}

The configuration space we establish based on the three principles covers all LATTE methods in~\Cref{tab:commands}. They are mapped onto the unified \LATTEArena{} pipeline according to our taxonomy, achieving unification beyond the 4 core dimensions in~\Cref{tab:revisiting}, thereby enabling a fair comparison of LATTE algorithms. We also provide detailed descriptions of the differences between original methods and the \LATTEArena{} implementation in~\Cref{tab:revisiting}.

\subsubsection{Remark on Uncovered Combinations}
We conduct repeated experiments for all methods, which is equivalent to combining each configuration with the Best-of-N strategy in LATTE. This will be reported as a performance metric ``Best'' (\Cref{tab:performance}) and is therefore not listed in~\Cref{tab:revisiting}. Additionally, all methods in \LATTEArena{} are tested under the AutoML setting, which incorporates model ensembling and effectively serves as a comparison for the Select-Expand-Ensemble strategy; for more details, refer to~\Cref{sssec:metric}.
The Rule output format restricts LLMs to binary feature operators for classification on linear models, demonstrating limited model and task adaptability, and is therefore excluded from our configuration space. 
We leave more complex configurations to future work, as they require novel techniques beyond the scope of \LATTEArena{}.

\section{Benchmarking and Findings}\label{sec:exp}
In this section, we evaluate LATTE techniques using \LATTEArena{}. We first specify the full experimental protocol, including dataset construction criteria, metric definitions, LLM backbones, and default hyperparameters. We then organize the empirical study around 7 core research questions (RQs), covering overall performance, cost, component effects, scalability, and robustness: \\
\textbf{RQs 1--3 (Overall Performance Comparison):}
\begin{itemize}[leftmargin=*, nosep]
    \item \textbf{RQ1:} What is the overall performance gain across diverse tabular classification and regression tasks? (\Cref{ssec:PGA})
    \item \textbf{RQ2:} What are the time and token overheads incurred by each LATTE configuration? (\Cref{ssec:TTCA})
    \item \textbf{RQ3:} How do configurations compare in token efficiency under \emph{fixed budgets}? (\Cref{ssec:TEA})
\end{itemize}
\textbf{RQs 4--5 (Component and Module Analysis):}
\begin{itemize}[leftmargin=*, nosep]
    \item \textbf{RQ4:} How do individual algorithmic components impact performance and cost? (\Cref{ssec:CLA})
    \item \textbf{RQ5:} How do peripheral pipeline modules (e.g., serializer) affect system efficacy? (\Cref{ssec:PMA})
\end{itemize}
\textbf{RQs 6--7 (Scalability and Robustness Analysis):} 
\begin{itemize}[leftmargin=*, nosep]
    \item \textbf{RQ6:} Do empirical findings generalize to \emph{large-scale} datasets? (\Cref{ssec:scale})
    \item \textbf{RQ7:} How do configuration choices influence pipeline viability and success rates? (\Cref{ssec:robust})
\end{itemize}

\subsection{Dataset}\label{ssec:data}
Existing LATTE research predominantly relies on classic benchmarks where SOTA tabular models already perform competitively without feature engineering, thus limiting the observable necessity and contribution of TAFE. To enable more rigorous and discriminative evaluation, we construct \LATTEArena{} using the ``hardest'' datasets from TabZilla~\cite{mcelfresh2023neural}, supplemented with datasets from an existing study~\cite{10.5555/3600270.3600307}. 

The construction adheres to three guiding principles rooted in LATTE's core motivation:
(1) Datasets should present non-trivial challenges, for which feature engineering is necessary and can yield meaningful performance gains. The specific selection criteria follow TabZilla's guidelines. Furthermore, we exclude datasets on which existing LATTE methods already achieve near-perfect performance, such as \textit{balance-scale} (\LLMFE{}~got 99\%) and \textit{car} (\LLMFE{}~got 100\%).
(2) Raw data and metadata are retained to preserve semantic fidelity and to enable LATTE's LLM backbone to utilize contextual information that traditional tabular models and TAFE methods typically cannot exploit.
(3) Features should exhibit heterogeneity, either across data types or in their underlying nature. For example, datasets dominated by homogeneous features, such as cnae-9~\cite{cnae-9_233} with word-frequency attributes, are excluded. This better reflects real-world TAFE application scenarios~\cite{10.5555/3600270.3600307}.

The final dataset suite spans multiple real-world domains, including healthcare, finance, and biology, covering sample sizes ranging from 294 to 1,025,009 instances and feature dimensions ranging from 5 to 118 features. The specific attributes of the datasets are listed in~\Cref{tab:dataset}.
To our knowledge, this is the first effort to define dataset selection criteria for the LATTE task. In \LATTEArena{}, we use these carefully curated datasets to evaluate existing technique combinations, addressing the lack of comparisons across different scales and tasks~\cite{hollmann2023large,lin2023smartfeat,han2024large}, while filtering out simple datasets to highlight performance gaps.

\begin{table}[htbp]\small
\centering
\caption{Dataset statistics. $N$ and $M$ count instances and features, respectively. Values in parentheses are the counts of numerical features. Kurtosis indicates the standard deviation of the kurtosis of all features. Large datasets (\mynote{}) are evaluated separately in~\Cref{ssec:scale}.}
\label{tab:dataset}
\renewcommand{\arraystretch}{1.0}
\begin{tabular}{lccccc}
\toprule
\multirow{2}{*}{\textbf{Datasets}} & \multicolumn{3}{c}{\textbf{Dataset Attributes}} & \multirow{2}{*}{\textbf{Metric}} & \multirow{2}{*}{\textbf{Test w/o FE}} \\
\cmidrule(lr){2-4} 
& N & M & Kurtosis\\
\midrule
heart-h        & 294 & 13 (5) & 3.88 & Accuracy & 78.65\\
credit-approval & 690 & 15 (6) & 55.19 & Accuracy & 88.05\\
vehicle        & 846 & 18 (18) & 15.16 & Accuracy & 72.94\\
credit-g        & 1,000 & 20 (3) & 5.58 & Accuracy & 73.83\\
qsar-biodeg        & 1,055 & 41 (32) & 127.32 & Accuracy & 87.21\\
socmob        & 1,156 & 5 (1) & 37.18 & Accuracy & 94.54\\
kc1        & 2,109 & 21 (21) & 29.17 & Accuracy & 85.59\\
nomao   & 34,465 & 118 (77) & 946.49 & Accuracy & 96.74\\
electricity   & 45,312 & 8 (8) & 2484.28 & Accuracy & 89.77\\
\mynote{}\textbf{ road-safety}   & 111,762 &  32 (32)  &  107.81  & Accuracy & 78.50\\
\mynote{}\textbf{ covertype}   & 423,680 &  55 (10)  &  1578.56  & Accuracy & 95.48\\
\mynote{}\textbf{ poker-hand}  & 1,025,009 &  10 (10)  &  0.08  & Accuracy & 73.88\\
\midrule
wine-quality & 6,487 & 11 (11) & 14.56 & N-RMSE & 0.1028\\
cpu-small & 8,192 & 12 (12) & 105.99 & N-RMSE & 0.0289\\
bike-sharing & 17,379 & 12 (7) & 8.92 & N-RMSE & 0.0461\\
diamonds & 53,940 & 9 (6) & 32.08 & N-RMSE & 0.0300\\
\bottomrule
\end{tabular}
\end{table}

\subsection{Evaluation Setting}\label{ssec:setting} 

\subsubsection{Metrics}\label{sssec:metric}
To comprehensively evaluate the performance of LATTE algorithms, the benchmark incorporates a set of complementary metrics covering predictive performance, computational efficiency, and robustness. 

\textbf{\emph{Performance Gain}} is evaluated using three metrics: VG (Validation Gain), TG (Test Gain), and AG (AutoML Test Gain). VG and TG measure absolute performance improvements on the validation and test sets using a fixed downstream model, with accuracy used for classification tasks and normalized root mean square error (N-RMSE) used for regression tasks. AG reflects the absolute performance improvements on the test set that LATTE brings to the AutoML pipeline. In our experiments, this is implemented using AutoGluon~\cite{agtabular}, a widely adopted tabular AutoML framework that integrates various models, including traditional methods such as Random Forest, gradient boosting trees such as LightGBM and XGBoost, as well as deep learning models such as NeuralNetFastAI.

\textbf{\emph{Computational Efficiency}} consists of two parts: token efficiency = $\frac{\text{Performance\ Gain}}{\text{Token Cost}}$ and time efficiency = $\frac{\text{Performance\ Gain}}{\text{Time Cost}}$. In \LATTEArena{}, the time cost primarily stems from LLM inference and evaluator training. The inference time is positively correlated with token cost, while the training time is affected by factors such as the downstream model size and computing hardware. Therefore, we focus more on the token cost.

\textbf{\emph{Robustness}} plays a critical role in LLM-driven data science research; however, it remains underexplored in most existing LATTE methods. To fill this gap, we perform repeated runs of each method on each dataset and compare their Success Rates.

\subsubsection{LLM Backbones and Parameters}
In this study, we select multiple LLMs for comparative experiments, covering both open-source and closed-source models, as well as thinking and non-thinking models. These include \texttt{GPT-4o}, \texttt{Deepseek-V3.1}, \texttt{o4-mini}, and \texttt{Llama-3.1-8B-Instruct}. The default LLM is \texttt{GPT-4o}, with temperature set to 1.

\LATTEArena{} employs a 3:1:1 split for training, validation, and testing, with experiments conducted over 6 random splits. The ToT family utilizes a binary tree with 5 exploration steps, while other families perform 10 iterations. The Evo family divides demonstrations into 5 populations. \Autofeat{}~runs for 2 steps. The default value for top-$k$ is 3, and the default downstream model is RandomForest. Following \ELLM{}, the warm-up module uses \GRFG{} and adopts its settings to collect 450 demonstrations. OPRO has 10 optimization steps in each iteration. Each run of a LATTE pipeline under a specific configuration and parameter setting is recorded as one execution log containing the prompt, LLM response, parsed operation, execution status, selected features, cost statistics, and downstream scores. Across all experiments, we preserve and publicly release over 4,000 logs to facilitate reproducibility, comparative analysis, and future community research.

\subsection{Performance Gain Analysis (\textbf{RQ 1})}\label{ssec:PGA}
The performance gains of different \LATTEArena{} configurations, together with the traditional TAFE method \Autofeat{}~\cite{horn2019autofeat} and RL-based TAFE method \GRFG{}~\cite{10.1145/3534678.3539278}, are shown in~\Cref{tab:performance}.

\emph{\textbf{Finding 1.} \uline{Task complexity dictates optimal search strategies, exposing a sharp divergence between classification and regression.}} In classification, computationally intensive, exploratory paradigms dominate, with Evo and OPRO families achieving the highest average Test Gains (TG) of 0.86\% (\EvoRPNw{}) and 0.73\% (\OPRORPN{}). Conversely, in regression, Greedy Incremental approaches exhibit superior efficiency: the CoT and ToT families secure the highest AutoML Test Gains (AG) (e.g., 0.10\% for \CoTCodeh{} and 0.08\% for \ToTRPN{}), whereas complex architectures like \OPROC{} fail to converge within fixed budgets and significantly underperform (-0.66\% AG). This may be because most existing works primarily design strategies and prompts targeting classification tasks, such as \FeatLLM{}, \LFG{}, and \OCTree{}.

\begin{table*}[htbp]\footnotesize
\centering
\caption{Performance gains across datasets (N-RMSE scaled by $\times$1{,}000; cold-started \EvoCode{}/\EvoRPN{} excluded as they fail to evolve within 10 iterations). The best metric value per family is \textbf{bolded}, with \colorbox{blue3}{top-3} and \colorbox{yellow3}{lowest} highlighted. \textbf{Reading Guide (\Cref{ssec:PGA}):} Observe how optimal strategies diverge between classification (Evo/OPRO) and regression (CoT/ToT) (\emph{Finding 1}); note the superiority of RPN/Code formats over NL (\emph{Finding 2}); note the severe LLM overfitting (e.g., \OPRORPN{}) via inflated Validation Gain (VG) vs.\ true downstream AutoML Test Gain (AG) (\emph{Finding 3}); and contrast zero-shot with history-based \texttt{h} variants, e.g., \ToTNL{} vs.\ \ToTNLh{} (\emph{Finding 4}).}
\label{tab:performance}
\setlength{\tabcolsep}{4pt}
\renewcommand{\arraystretch}{1.2}
\resizebox{\textwidth}{!}{%
\begin{tabular}{ccIcIcIcIcccccccccIccccccIcccIcccIc}
\toprule
& & & & & \multicolumn{9}{cI}{\textbf{CoT}} & \multicolumn{6}{cI}{\textbf{ToT}} & \multicolumn{3}{cI}{\textbf{Critic}} & \multicolumn{3}{cI}{\textbf{OPRO}} & \textbf{Evo} \\
\cline{6-27}
& & & \textcolor{gray}{\Autofeat{}} & \textcolor{gray}{\GRFG{}} & \CoTNL{} & \CoTNLh{} & \CoTNLt{} & \CoTCode{} & \CoTCodeh{} & \CoTCodet{} & \CoTRPN{} & \CoTRPNh{} & \CoTRPNt{} & \ToTNL{} & \ToTNLh{} & \ToTCode{} & \ToTCodeh{} & \ToTRPN{} & \ToTRPNh{} & \CriticNL{} & \CriticCode{} & \CriticRPN{} & \OPROC{} & \OPRO{} & \OPRORPN{} & \EvoRPNw{} \\
\midrule
\multirow{6}{*}{\rotatebox{90}{\textbf{Classification}}} 
& \multirow{3}{*}{\rotatebox{90}{Average}} 
& VG & \textcolor{gray}{0.31} & \textcolor{gray}{3.02} & 2.24 & 2.64 & 2.48 & 2.70 & \cellcolor{yellow3}2.04 & 2.20 & \textbf{2.74} & 2.64 & 2.60 & 2.46 & 2.43 & 2.42 & 2.31& \textbf{2.81} & 2.57 &2.30 &2.17 & \textbf{2.57} & \cellcolor{blue3}3.20 &3.13 & \cellcolor{blue1}\textbf{3.55} & \cellcolor{blue2}3.23 \\
& & TG & \textcolor{gray}{0.72} & \textcolor{gray}{0.56} & 0.39 & 0.04 & 0.44 & 0.20 & 0.45 & 0.34 & \textbf{0.48} & 0.25 & -0.05 & 0.25 & 0.28 & 0.24& \cellcolor{yellow3}-0.06 & \textbf{0.42} & 0.41 &0.17 &\textbf{0.35} & 0.08 & \cellcolor{blue3}0.71 &0.56 &\cellcolor{blue2}\textbf{0.73} & \cellcolor{blue1}0.86 \\
& & AG & \textcolor{gray}{0.31} & \textcolor{gray}{-0.07} & -0.36 & 0.00 & 0.05 & \cellcolor{yellow3}-0.69 & -0.34 & -0.50 & \cellcolor{blue3}\textbf{0.09} & -0.28 & -0.20 & -0.64 & \cellcolor{blue3}\textbf{0.09} & -0.73 &-0.45 & -0.42 & -0.05 &\cellcolor{blue2}0.12 &\cellcolor{blue1}\textbf{0.13} & -0.48 & -0.54 &\textbf{0.08} &-0.32 & \cellcolor{blue2}0.12 \\
\cmidrule(lr){2-27}
& \multirow{3}{*}{\rotatebox{90}{Best}} 
& VG & \textcolor{gray}{1.08} & \textcolor{gray}{4.63} & \cellcolor{yellow3}3.33 & 4.13 & 3.62 & 3.97 & 3.38 & 3.40 & 3.72 &\cellcolor{blue2}\textbf{4.78} & 3.53 & 3.80 & 3.85 & 3.48&3.34 & 3.80 & \textbf{3.86} &3.84 & 3.55& \textbf{4.42} & \cellcolor{blue3}4.66 & 4.28 & \cellcolor{blue1}\textbf{5.15} & 4.38 \\
& & TG & \textcolor{gray}{2.23} & \textcolor{gray}{2.71} & 1.91 & 2.03 & \cellcolor{blue3}\textbf{2.69} & 1.90 & 1.91 & 1.77 & 1.97 & 2.15 & 1.75 & 1.86 & 2.04 &1.66 &1.56 & 2.37 & \cellcolor{blue2}\textbf{2.74} & \cellcolor{yellow3}1.52 &\textbf{2.11} & 1.56 & \cellcolor{blue1}\textbf{2.79} & 1.99 & 2.45 & \cellcolor{blue1}2.79 \\
& & AG & \textcolor{gray}{1.35} & \textcolor{gray}{1.73} & 1.01 & 1.83 & 1.88 & \cellcolor{yellow3}0.58 & 0.85 & 0.93 & 2.21 & 1.24 & \cellcolor{blue3}\textbf{2.28} & 1.01 & \cellcolor{blue1}\textbf{2.42} &0.74 &1.38 & 1.44 & 2.16 &1.65 &1.89 & \textbf{2.01} & 0.67 &1.67 & \textbf{1.74}& \cellcolor{blue2}2.32 \\
\midrule
\multirow{6}{*}{\rotatebox{90}{\textbf{Regression}}} 
& \multirow{3}{*}{\rotatebox{90}{Average}} 
& VG & \textcolor{gray}{\textless 0} & \textcolor{gray}{0.57} &  0.48 & 0.70 & 0.45 & 0.83 & \textbf{0.84} &0.67 & 0.81 & 0.56 & 0.68 & 0.64 & \cellcolor{yellow3}0.43 & \textbf{0.78}& 0.73& 0.54 & \textbf{0.78} &0.52 &\textbf{0.68} & 0.64 & \cellcolor{blue3}1.02 &\cellcolor{blue1}\textbf{1.71} &\cellcolor{blue2}1.51 & 0.50 \\
& & TG & \textcolor{gray}{--} & \textcolor{gray}{0.42} & 0.43 & 0.05 & 0.46 & \cellcolor{blue1}\textbf{0.73} & \cellcolor{blue3}0.59 & 0.52& 0.35 & 0.18 & 0.26 & 0.10 & \cellcolor{yellow3}-0.01 & \textbf{0.54}&0.22 & 0.32 & 0.48 &0.44 &\textbf{0.46} & 0.40 & \cellcolor{blue1}\textbf{0.73} & \cellcolor{blue2}0.72 &0.34 & 0.23 \\
& & AG & \textcolor{gray}{--} & \textcolor{gray}{-0.18} & 0.04 & -0.26 & -0.11 & 0.05 & \cellcolor{blue1}\textbf{0.10} & \cellcolor{blue3}0.07& \cellcolor{blue2}0.08 & -0.21 & -0.02 & -0.11 & 0.02 & -0.06&\cellcolor{blue3}0.07 & \cellcolor{blue2}\textbf{0.08} & -0.05 &\textbf{0.00} &-0.47 & -0.28 & -0.66 & \textbf{-0.41} & \cellcolor{yellow3}-0.85 & -0.62 \\
\cmidrule(lr){2-27}
& \multirow{3}{*}{\rotatebox{90}{Best}} 
& VG & \textcolor{gray}{\textless 0} & \textcolor{gray}{0.61} & 0.99 & 1.01 & 0.81 & 1.05 & 1.04 & 0.87& \cellcolor{blue3}\textbf{1.29} & 0.90 & 0.79 & 0.92 & \cellcolor{yellow3}0.61 & 0.85& 0.94 & 0.68 & \textbf{1.09} & 0.67&\textbf{0.95} & 0.80 & 1.03 &\cellcolor{blue1}\textbf{1.91} & \cellcolor{blue2}1.90 & \cellcolor{yellow3}0.61 \\
& & TG & \textcolor{gray}{--} & \textcolor{gray}{0.46} & 0.81 & \cellcolor{yellow3}0.30 & 0.52 & \cellcolor{blue2}\textbf{1.20} & \cellcolor{blue3}1.02 &0.70 & 0.79 & 0.37 & 0.52 & 0.41 & 0.32 &0.99 &0.43 & 0.49 & \textbf{1.00} &0.75 &\textbf{0.95} & 0.75 & 0.78 &\cellcolor{blue1}\textbf{1.23} & 0.93 & 0.39 \\
& & AG & \textcolor{gray}{--} & \textcolor{gray}{-0.14} & 0.21 & -0.04 & 0.18 & \cellcolor{blue2}0.22 & \cellcolor{blue1}\textbf{0.23} &0.19 & 0.13 & 0.04 & \cellcolor{blue3}0.21 & -0.03 & \textbf{0.21} &0.12 &0.19 & 0.17 & 0.16 &0.07 & -0.02& \textbf{0.16} & \cellcolor{yellow3}-0.63 & \textbf{-0.15} & -0.17 & -0.09 \\
\bottomrule
\end{tabular}%
}
\end{table*}

\emph{\textbf{Finding 2.} \uline{Structured formats (RPN and Code) systematically overcome the expressive bottleneck of NL.}} Examining the distribution of best performance reveals clear trends: within the CoT family, \CoTCode{} and its variants lead in regression tasks, while \CoTRPN{} and its variants lead in classification tasks; similar patterns can be observed in the ToT family, the Critic family, and the OPRO family. Through analysis of LLM outputs, this phenomenon can be attributed to two reasons: (1) As discussed in \Cref{ssec:format}, RPN and Code formats can express high-order complex features, expanding the exploration space and thus outperforming NL; (2) Code has a lower execution success rate due to its inherent complexity (see \Cref{ssec:robust}), making it slightly inferior to RPN. However, on regression datasets, we observe that many high-reward FE behaviors in Code outputs rely on row-wise processing, such as group-by-then-mapping operations, which involve aggregation operators and complex combinations that are challenging for LLM reasoning in RPN. Additionally, Code can perform operations like data selection, exceeding RPN's expressive capability.

\emph{\textbf{Finding 3.} \uline{The evaluators in LATTE methods exhibit severe overfitting, necessitating robust sampling strategies.}} This reflects a fundamental limitation in current LATTE methods: the Evaluator computes scores using the validation model, and both prompts and strategies are optimized against these scores. With small validation sets, the LLM outputs become overly tailored to specific data distributions and models, showing a declining trend from VG to TG and AG (e.g., \OPRORPN{} reports 3.55\% average VG but only 0.73\% TG and -0.32\% AG in classification). Best-of-N sampling (e.g., \EvoRPNw{}) partially mitigates this performance collapse by diversifying candidate populations before selection, elevating AG to a positive 0.12\%.

\emph{\textbf{Finding 4.} \uline{Naive history demonstrations degrade search trajectories.}} As shown in~\Cref{tab:performance}, equipping CoT or ToT with history demonstrations (\texttt{h} variants) fails to uniformly improve upon zero-shot counterparts, often causing VG and TG degradation (e.g., \CoTRPN{} vs.\ \CoTRPNh{} and \ToTCode{} vs.\ \ToTCodeh{}). Existing demonstrations report only newly added features and their scores, omitting the full evolution of the feature set and performance. As a result, context-independent marginal scores isolate features from holistic interactions, confusing the LLM evaluator and misguiding the evolutionary direction.

\subsection{Time and Token Cost Analysis (\textbf{RQ 2})}\label{ssec:TTCA}
\begin{figure*}
    \centering
    \includegraphics[width=1.0\linewidth]{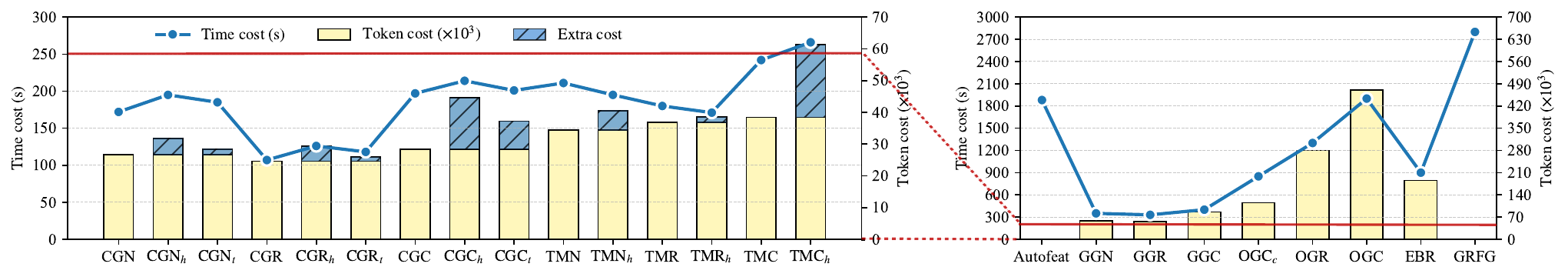}
\caption{Average time/token cost (left/right y-axis scale ratio is 1:10; shaded areas show demonstration overhead). \textbf{Reading Guide (\Cref{ssec:TTCA}):} Observe the exponential cost surge from CoT to iterative OPRO/Evo (\emph{Finding 5}); LATTE's efficiency over traditional \Autofeat{} in terms of time cost (\emph{Finding 6}); and how demonstrations (blue shaded) and Code format (\CoTNL{} vs.\ \CoTCode{}) primarily inflate input context (\emph{Finding 7}).}
    \label{fig:cost}
\end{figure*}

Beyond performance gains, the practical deployment of LATTE methods also depends on their cost. We analyze both token and time usage to evaluate the cost-effectiveness of different configurations. \Autofeat{} and \GRFG{} do not use LLM and thus incur no token cost. For \EvoRPNw{}, the overhead of the evolutionary part is recorded in \EvoRPN{}, while the warm-up overhead is recorded in \GRFG{}. The average cost of each method is shown in~\Cref{fig:cost}.

\emph{\textbf{Finding 5.} \uline{Architectural complexity drives exponential, rather than linear, increases in time and token overheads.}}
We partition configurations into two groups based on overhead, shown in the left and right panels of~\Cref{fig:cost}, respectively. The low-cost group includes the CoT and ToT families, while the high-cost group comprises Critic, OPRO, and Evo. While upgrading from single-pass CoT to ToT or Critic architectures yields moderate overhead growth (1.3$\times$ and 2$\times$, respectively), highly iterative frameworks (Evo and OPRO) trigger a $\sim$10$\times$ surge in token and time costs, exposing a steep scalability cliff.

\emph{\textbf{Finding 6.} \uline{LATTE methods establish a new time-efficient paradigm compared to traditional RL-based algorithms.}} 
Most LATTE variants require only about 10\% of the runtime of \Autofeat{} or \GRFG{}, with \CoTRPN{} as low as 5\%. Even highly iterative models (OPRO-based, \EvoRPNw{}) consume only 30--60\% of baseline runtime despite multiple LLM queries per iteration and additional validation model training or evolution steps. This speedup stems from the LLMs' semantic-driven exploration, which drastically reduces the exhaustive candidate evaluations required by traditional heuristic loops.

\emph{\textbf{Finding 7.} \uline{Historical demonstrations inflate input context, whereas output format governs generation latency.}}
We highlight with shading in~\Cref{fig:cost} the extra cost that demonstrations bring to each base configuration, and report detailed analysis in~\Cref{tab:ablation_cost}. Demonstrations primarily inflate input context rather than altering structural outputs. By contrast, output format governs generation latency: RPN cuts time and output tokens, whereas Code increases both. When Code is used as the output format, the flexible number and complexity of operations cause substantial variation in LLM output length, typically about 2.5$\times$ that of the NL format. This flexibility makes it difficult to compress historical records, leading to a significant increase in demonstration length.

\begin{figure*}
    \centering
    \includegraphics[width=1.0\linewidth]{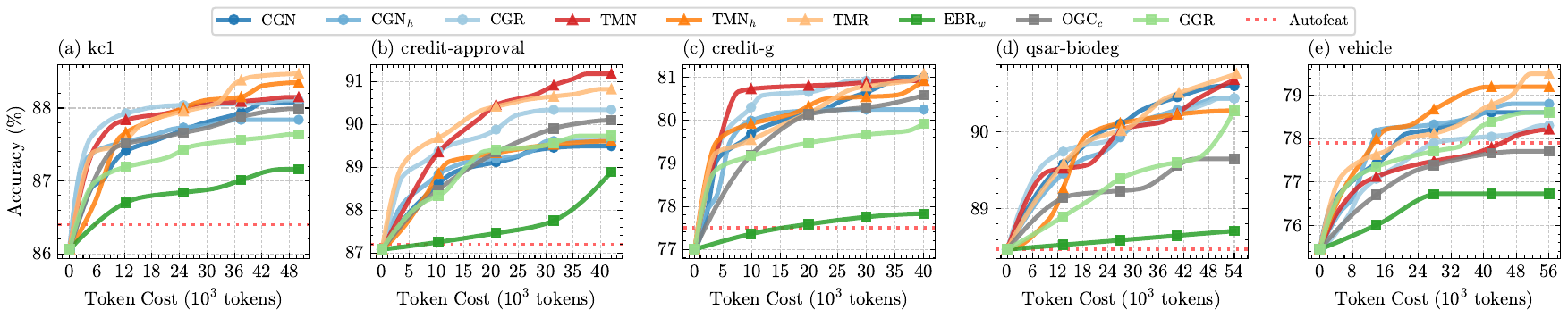}
\caption{Accuracy vs. Token Cost for 9 representative methods on 5 classification datasets. \textbf{Reading Guide (\Cref{ssec:TEA}):} Note the crossover between CoT's early convergence ($\bullet$) and ToT's sustained scaling ($\scriptstyle\blacktriangle$) (\emph{Finding 8}); the prohibitive cost thresholds of complex methods like Evo and OPRO ($\scriptstyle\blacksquare$) (\emph{Finding 9}); and how historical demonstrations inflate token costs without proportional gains (\emph{Finding 10}).}
    \label{fig:cost2}
\end{figure*}

\begin{figure*}
    \centering
    \includegraphics[width=1.0\linewidth]{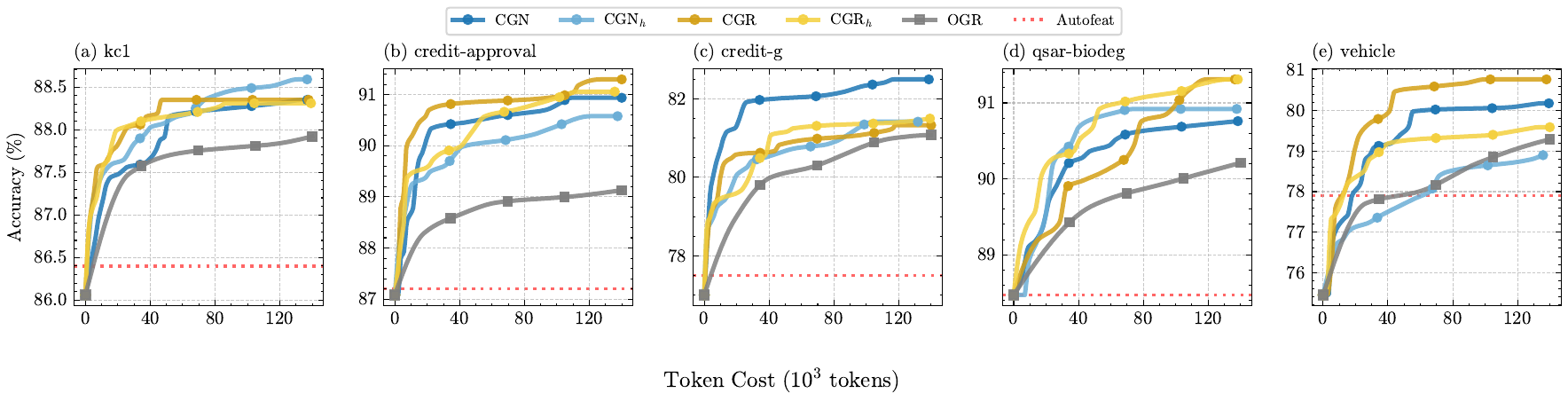}
    \caption{VG vs. Token Cost for simple CoT methods and \OPRORPN{}~with high budgets. \textbf{Reading Guide (\Cref{ssec:TEA}):} Note how zero-shot CoT overtakes \OPRORPN{} once given an equal token budget (\emph{Extended Finding}).}
    \label{fig:cost3}
\end{figure*}

\subsection{Token Efficiency Analysis (\textbf{RQ 3})}\label{ssec:TEA}
Previous studies have lacked a systematic comparison of different LATTE methods under the same token budget, limiting practical guidance for real-world applications. From ~\Cref{fig:cost} and ~\Cref{tab:performance}, we observe that low-cost CoT methods have already achieved leading performance on regression tasks, but the cost-effectiveness of different methods on classification tasks is still difficult to compare intuitively. Therefore, in this section, we select 9 methods and plot VG curves against token consumption in~\Cref{fig:cost2}. Methods are selected based on their cost-effectiveness under fixed-round settings. Datasets are selected to exhibit large VG variance across methods, which facilitates comparison analysis.
\emph{\textbf{Finding 8.} \uline{CoT methods excel in low-budget scenarios, while ToT methods scale robustly across extended budgets.}} Our analysis reveals that CoT methods exhibit impressive initial VG growth but quickly reach local optima due to their greedy strategies. In contrast, ToT methods, though slightly underperforming compared to the best CoT variants at low budgets, consistently surpass them in high-budget scenarios by balancing exploration and exploitation.

\emph{\textbf{Finding 9.} \uline{Complex architectures require massive token investments to overcome search overhead.}} Despite potential performance gains, Critic, OPRO, and Evo family methods exhibit poor token efficiency under tight budgets. For instance, the early-stage lag of \EvoRPNw{} proves their advantages strictly depend on relaxed cost constraints, thus limiting practical deployment.

\emph{\textbf{Finding 10.} \uline{Demonstrations severely dilute token efficiency, eclipsing their marginal semantic gains.}} Although Positive-Negative demonstrations (those \texttt{h} variants) may enhance absolute performance in fixed-round settings, these slight gains are entirely offset by the compounded input token overhead per LLM call. Consequently, zero-shot structured prompts remain fundamentally more cost-effective.

To further investigate the cost-effectiveness of different LATTE configurations, we extend the CoT budget until its cumulative token consumption reaches the same order as \OPRORPN{}, the best-performing high-cost method under fixed-round settings (140k+ tokens). This experiment asks whether OPRO's sophisticated iterative refinement remains advantageous once simple methods are granted the same total query budget. The results are shown in~\Cref{fig:cost3}. We observe that simple CoT methods, even those without demonstrations, consistently outperform \OPRORPN{}.

\emph{\textbf{Extended Finding.} \uline{OPRO iteratively optimizes the quality of a single output, which is less cost-effective than multiple independent LLM queries under high-budget CoT scaling.}}

\subsection{Component-Level Analysis (\textbf{RQ 4})}\label{ssec:CLA}

\subsubsection{Analysis of CoT and ToT Family} 
To further investigate the effects of different configurations on performance and cost, we conducted a detailed analysis of CoT and ToT variants (\Cref{tab:ablation_cost}).

\emph{\uline{Unlike CoT, some ToT variants exhibit an inverse time-token relationship driven by MCTS dynamics.}}
As shown in~\Cref{fig:cost}, the ToT family displays patterns that diverge from CoT, particularly in time cost trends. Statistics for these exceptional variants in~\Cref{tab:ablation_cost} reveal that MCTS-induced variability in LLM query counts drives this inverse relationship: (1) When an optimal path is apparent, MCTS focuses on exploitation and visits fewer nodes. Providing demonstrations enhances the model's ability to identify promising paths early, resulting in fewer queries with lower total time, despite higher per-query token consumption. (2) When high-value nodes appear at varying depths, the NL format hinders their discovery. The RPN format, by enabling the generation of high-order features, brings high-value nodes to shallower tree levels, allowing MCTS to discover more nodes with comparable UCB values and thereby promoting exploration. While this increases the number of queries, per-query latency significantly decreases (a pattern observed in the CoT family), ultimately reducing overall time.

\begin{table}[htbp]\small
\setlength{\tabcolsep}{6pt}
\centering
\caption{Cost changes of CoT/ToT configurations (normalized to 100\% via ``---''). \textbf{Reading Guide (\Cref{ssec:TTCA}):} Note format-driven latency (\emph{Finding 7}): RPN cuts time and output tokens, whereas Code increases both. Adding demonstrations primarily inflates input context rather than altering structural outputs.}
\label{tab:ablation_cost}
\renewcommand{\arraystretch}{1.2}
\begin{tabular}{lccc}
\toprule
\textbf{CoT family} & $\Delta$\textbf{ Time (\%)} & $\Delta$\textbf{ Token (\%)} & $\Delta$\textbf{ Output Token (\%)}\\
\midrule
\CoTNL{} & --- & --- & ---\\
+ Positive–Negative & $\uparrow 13.4$ & $\uparrow 19.6$ & $\uparrow 45.7$ \\
+ Top-k & $\uparrow 7.6$ & $\uparrow 6.8$ & $\uparrow 53.1$\\
- NL + RPN & $\downarrow 36.1$ & $\downarrow 7.5$ & $\downarrow 81.3$\\
- NL + RPN + Positive–Negative & $\downarrow 24.5$ & $\uparrow 9.9$ & $\downarrow 78.7$\\
- NL + Code & $\uparrow 11.3$ & $\uparrow 6.8$ & $\uparrow \sim150$ \\
- NL + Code + Positive–Negative & $\uparrow 24.4$ & $\uparrow 67.7$ & $\uparrow \sim150$\\
\midrule
\textbf{ToT family} & $\Delta$\textbf{ Time (\%)} & $\Delta$\textbf{ Token (\%)} & \textbf{AVG Query Num}\\
\midrule
\ToTNL{} & --- & --- & 13.1\\
+ Positive–Negative & $\downarrow 7.6$ & $\uparrow 17.6$ & 12.4\\
- NL + RPN & $\downarrow 15.7$ & $\uparrow 6.9$ & 13.8\\
- NL + RPN + Positive–Negative & $\downarrow 19.1$ & $\uparrow 11.7$ & 13.3\\
\bottomrule
\end{tabular}
\end{table}

\subsubsection{Analysis of OPRO and Evo Family} 
Additionally, we conducted extended experiments (\Cref{tab:extend}) to analyze how technique components in OPRO- and Evo-family methods affect token efficiency.
Following \OCTree{}, \OPROC{} integrates OPRO with CART-based reasoning. To evaluate the impact of each technique, we performed ablation studies by removing them individually.
We found that:
(1) CART is a cost-effective substitute for metadata when using code output format, substantially reducing token cost while maintaining performance.
(2) Although OPRO yields substantial performance improvements, the associated 19$\times$ increase in token costs severely compromises the method's overall cost-effectiveness.

\emph{\textbf{Finding 11.} \uline{Lightweight structural priors (CART) deliver massive cost savings, whereas the iterative OPRO refinement loop monopolizes the overhead budget.}} Ablating CART reasoning in favor of raw dataset metadata inflates token consumption by $\sim$280\% while slightly degrading VG ($\downarrow$2.2\%), proving tree-based priors are a highly cost-effective surrogate for injecting raw statistical prompts. Conversely, while the OPRO feedback loop drives substantial test gains (TG drops by 66.2\% without it), it triggers an extreme token surge (indicated by a $\sim$95\% cost drop when removed). This renders the full iterative pipeline impractical for standard tabular tasks unless budget constraints are entirely relaxed.

For the Evo family, considering that \EvoRPN{} and \EvoCode{} require consuming a large number of tokens (\textgreater 100k) during the cold start phase, we use \EvoRPNr{} to replace them. It uses a random RPN collector for warm-up, thereby avoiding the additional token cost. \Cref{tab:extend} shows that (1) \EvoRPNw{}'s evolutionary phase predominantly enhances TG and AG, indicating the LLM's ability to leverage semantic information for generating features with superior generalization. (2) The data collector has a critical impact on the final performance.

\emph{\textbf{Finding 12.} \uline{Evolutionary mutation synthesizes generalizable features, but warm-up collector quality dictates the performance ceiling.}} The evolutionary mutation phase strongly boosts downstream generalization, with TG and AG plummeting by 34.9\% and 158\% when reverting strictly to \GRFG{}. This confirms that the LLM leverages semantic information for robust feature synthesis rather than merely overfitting the validation set. However, this gain is strictly gated by initialization: replacing the \GRFG{} warm-up with a random collector collapses AG by 358\%, showing that the initial collector quality, not the mutation operator itself, is the decisive performance bottleneck.

\begin{table}[htbp]\small
\setlength{\tabcolsep}{5pt}
\centering
\caption{Performance and token cost evaluation for OPRO/Evo. \textbf{Reading Guide (\Cref{ssec:CLA}):} Note CART's cost-efficient surrogacy versus the OPRO loop's token surge (\emph{Finding 11}), and the \GRFG{} warm-up collector's grip on Evo's gains (\emph{Finding 12}).}
\label{tab:extend}
\renewcommand{\arraystretch}{1.2}
\begin{tabular}{lcccc}
\toprule
\textbf{OPRO family} &$\Delta$ \textbf{ VG (\%)} & $\Delta$\textbf{ TG (\%)} & $\Delta$\textbf{ AG (\%)} & $\Delta$\textbf{ Token (\%)}\\
\midrule
\OPROC{} & --- & --- & ---& ---\\
- CART + Metadata (\OPRO{}) & $\downarrow 2.2$ & $\downarrow 1.4$& $\uparrow \sim50$ & $\uparrow \sim280$\\
- OPRO (\CoTCodec{}) & $\downarrow 11.9$ & $\downarrow 66.2$ & $\uparrow 44.4$ & $\downarrow \sim95 $\\
- OPRO - CART + Metadata (\CoTCode{}) & $\downarrow 16.6$ & $\downarrow 71.8$ & $\downarrow 27.8$ & $\downarrow \sim75$\\
\midrule
\textbf{Evo family} & $\Delta$\textbf{ VG (\%)} & $\Delta$\textbf{ TG (\%)} & $\Delta$\textbf{ AG (\%)} &  $\Delta$\textbf{ Token (\%)}\\
\midrule
\EvoRPNw{} & --- & --- & ---& ---\\
- EvoPrompt (\GRFG{}) & $\downarrow 6.5$ & $\downarrow 34.9$ & $\downarrow 158$ & $\downarrow 100$\\
- GRFG + Random Collector (\EvoRPNr{}) & $\downarrow 9.9$ & $\downarrow 25.6$ & $\downarrow 358$ & $\sim0.0$\\
\bottomrule
\end{tabular}
\end{table}

\subsubsection{Analysis of Three Configuration Principles} 
In ~\Cref{ssec:revisit}, we proposed three configuration principles for combining existing techniques. In this subsection, we conduct quantitative experiments to validate these principles, with the results presented in ~\Cref{tab:CP}. As observed, combining ToT with a greedy strategy and using NL as the output format in the OPRO family both yield poor performance, which is consistent with the second principle \emph{respecting component constraints}. Additionally, Least-to-Most (LtM) prompting improves performance on CoT without history but degrades it on CoT with history. Since LtM provides no overall performance gain while increasing overhead by 3 to 5 times, it is filtered out by the third principle \emph{prioritizing cost-effectiveness}.

\begin{table}[htbp]\small
\setlength{\tabcolsep}{7pt}
\centering
\caption{Quantitative validation of Configuration Principles. \textbf{Reading Guide (\Cref{ssec:CLA}):} Note how mismatched components (ToT+Greedy, OPRO+NL) degrade VG (\emph{Principle 2}), while LtM inflates cost 3--5$\times$ without gain (\emph{Principle 3}).}
\label{tab:CP}
\renewcommand{\arraystretch}{1.2}
\begin{tabular}{lccc}
\toprule
\textbf{ToT family} & $\Delta$\textbf{ VG (\%)} &  $\Delta$\textbf{ Token (\%)} &  $\Delta$\textbf{ Time (\%)}\\
\midrule
- MCTS + Greedy & $\downarrow 6.0$ & $\uparrow 3.9$ & $\downarrow 13.7$\\
\midrule
\textbf{OPRO family} & $\Delta$\textbf{ VG (\%)} &  $\Delta$\textbf{ Token (\%)} &  $\Delta$\textbf{ Time (\%)}\\
\midrule
- RPN + NL & $\downarrow 14.4$  & $\uparrow 10.7$ & $\uparrow \sim140$\\
- Code + NL & $\downarrow 6.8$ & $\downarrow 22.3$ & $\downarrow 40.4$\\
\midrule
\textbf{CoT family} &$\Delta$ \textbf{ VG (\%)} & $\Delta$\textbf{ Token (\%)} &  $\Delta$\textbf{ Time (\%)}\\
\midrule
+ LtM (CoT w/o history) & $\uparrow 11.8$ & $\uparrow \sim430$ & $\uparrow \sim430$\\
+ LtM (CoT w history) & $\downarrow 11.8$ & $\uparrow \sim570$ & $\uparrow \sim230$\\
+ LtM (all CoT baseline) & $\downarrow 0.4$ & $\uparrow \sim500$ & $\uparrow \sim320$\\
\bottomrule
\end{tabular}
\end{table}

\subsection{Pipeline Module Analysis (\textbf{RQ 5})}\label{ssec:PMA}

So far, we have conducted a detailed comparison of the four core dimensions in the six-dimensional taxonomy. For the remaining two dimensions, we employed a uniform experimental setup (\Cref{sssec:config}). To validate the rationality of our default settings, we conducted extended experiments on the corresponding modules: the Serializer, the Feature Selector, and the LLM. 

\begin{table}[htbp]\small
\centering
\caption{Ablation of the serializer and feature selector. \textbf{Reading Guide (\Cref{ssec:PMA}):} Observe the severe success rate drop without metadata and samples (\emph{Component Synergy}), the token savings from omitting calculated values (\emph{Informational Redundancy}), and the feature selector's massive impact on latency and VG (\emph{Pipeline Bottlenecks}).}
\setlength{\tabcolsep}{5pt} 
\renewcommand{\arraystretch}{1.2}
\begin{tabular}{lcccc}
\toprule
\textbf{Method} & $\Delta$\textbf{Success Rate (\%)} & $\Delta$\textbf{VG (\%)} & $\Delta$\textbf{Token (\%)} & $\Delta$\textbf{Time (\%)}\\
\midrule
\LATTEArena{} (\CoTNL{} + \CoTNLh{} + \CoTRPN{}) & --- & --- & --- & ---\\
- Calculated Values & $\downarrow 6.0$ & $\downarrow 2.1$ & $\downarrow 33.1$ & $\downarrow 24.0$ \\
- Metadata & $\downarrow 9.6$ &  $\downarrow 6.4$ & $\downarrow 38.9$ & $\downarrow 26.8$ \\
- (Data) Samples & $\downarrow 2.4$ &  $\downarrow 3.0$ & $\downarrow 21.6$ & $\downarrow 8.9$ \\
- Metadata \& Samples & $\downarrow 16.5$ & $\downarrow 16.8$ & $\downarrow 75.7$ & $\downarrow 34.4$ \\
- Feature Selector & $\sim 0.0$ & $\downarrow 39.5$ & --- & $\downarrow 48.1$ \\
+ Generated Metadata & $\downarrow 3.6$  & $\uparrow 14.6$ & $\uparrow\ \sim40$ & $\uparrow\ \sim100 $ \\
\bottomrule
\end{tabular}
\label{tab:ablation}
\end{table}

\subsubsection{Analysis on Serializer and Feature Selector}

\Cref{tab:ablation} shows the ablation results of the serializer and feature selector. For generated metadata, the LLM rewrites it using the native metadata, calculated values, and data samples. When removing metadata, we retain concise feature names that contain semantic information.
\begin{itemize}[leftmargin=*]
\item \emph{\uline{Component Necessity and Synergy.}} Any component removal degrades the success rate and VG, confirming the rationale of the \LATTEArena{} pipeline design. Specifically, the simultaneous exclusion of Metadata and Samples results in the largest drop in success rate, demonstrating their synergistic role in LATTE.
\item \emph{\uline{Informational Redundancy.}} Removing calculated values, metadata, or data samples slashes token costs by over 20\% but only yields marginal VG losses. This disproportion suggests significant informational redundancy in tables and metadata for LATTE.
\item \emph{\uline{Pipeline Bottlenecks.}} Removing the selector reduces latency by 48.1\% but causes a catastrophic VG collapse. Generating metadata achieves higher VG, suggesting room for optimization in the metadata format, though at the cost of significant overhead.
\end{itemize}
\emph{\textbf{Finding 13.} \uline{Pipeline bottlenecks dictate optimization priorities: Feature selection acts as the primary lever for balancing temporal efficiency against predictive performance, whereas metadata management (generation vs. compression) directly governs the token-accuracy trade-off.}}

\begin{figure}
  \centering
  \includegraphics[width=0.8\linewidth]{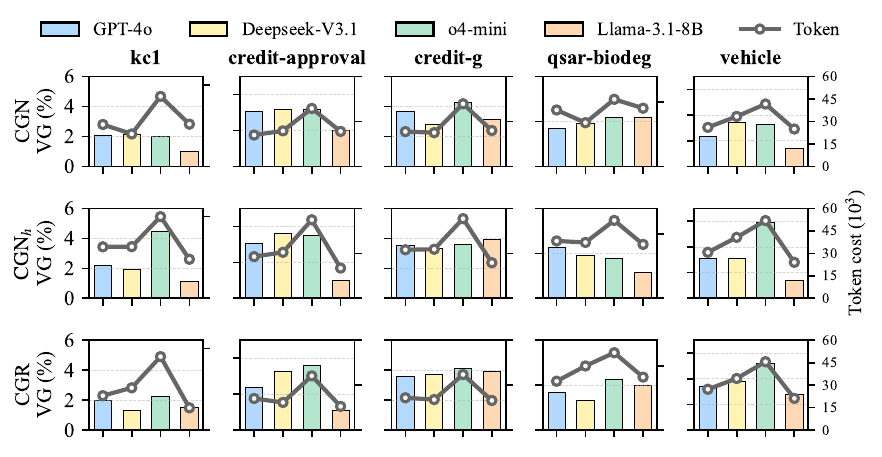}
  \caption{Performance and cost of different LLM backbones. \textbf{Reading Guide (\Cref{ssec:PMA}):} \texttt{Deepseek-V3.1} achieves Pareto optimality comparable to \texttt{GPT-4o}; \texttt{o4-mini} secures the highest VG but incurs an 80\% token premium; \texttt{Llama-3.1-8B} struggles with instruction-following formatting bottlenecks.}
  \label{fig:llm}
\end{figure}

\subsubsection{Analysis on LLM Backbones}\label{sssec:llms}
Experiments show that the best LATTE method with \texttt{GPT-4o} outperforms both traditional and RL-based TAFE methods in efficiency and performance. However, since the quality of FE operations depends on the LLM's reasoning capability, a natural question arises: how do LATTE methods perform with LLMs of varying capabilities and scales? \emph{\LATTEArena{} provides a convenient tool for assessing the data engineering capabilities of various LLMs and offers an additional dimension for LLM evaluation.}

We evaluated \LATTEArena{} using \texttt{Deepseek-V3.1}, \texttt{o4-mini}, and \texttt{Llama-3.1-8B-Instruct} as representative LLMs. \Cref{fig:llm} shows the VG and token cost of different LLMs.
\texttt{Deepseek-V3.1} achieves performance and cost most comparable to \texttt{GPT-4o}. \texttt{o4-mini} achieves the highest VG but its reasoning process incurs 80\% additional token cost compared to \texttt{GPT-4o}. \texttt{Llama-3.1-8B-Instruct}, as the smallest model, exhibits the poorest performance. It struggles to discover effective features across most datasets and demonstrates weak instruction-following capabilities, frequently failing to produce responses that adhere to the required format and feature operation rules. Moreover, its NL reasoning outputs tend to be brief, resulting in minimal token overhead. This indicates that the LATTE task capabilities on smaller models still require further enhancement through training techniques such as fine-tuning.

\begin{table*}[htbp]\footnotesize
\centering
\caption{Performance on large datasets. ``r \& c'' averages \emph{road-safety} and \emph{covertype} datasets; the \emph{poker-hand} (``poker'') dataset is isolated due to differing gain magnitudes. \textbf{Reading Guide (\Cref{ssec:scale}):} Note the narrowed VG-TG gap (\emph{Finding 14}), and the Code format's absolute dominance on complex-logic tasks like poker-hand (\emph{Finding 15}).}
\label{tab:scale}
\setlength{\tabcolsep}{4pt} 
\renewcommand{\arraystretch}{1.2}
\resizebox{\textwidth}{!}{%
\begin{tabular}{cIcIcIcIcccccccccIccccccIcccIccccIc}
\toprule
& & & & \multicolumn{9}{cI}{\textbf{CoT}} & \multicolumn{6}{cI}{\textbf{ToT}} & \multicolumn{3}{cI}{\textbf{Critic}} & \multicolumn{4}{cI}{\textbf{OPRO}} & \textbf{Evo} \\
\cline{5-27}
& & \textcolor{gray}{\Autofeat{}} & \textcolor{gray}{\GRFG{}} & \CoTNL{} & \CoTNLh{} & \CoTNLt{} & \CoTCode{} & \CoTCodeh{} & \CoTCodet{} & \CoTRPN{} & \CoTRPNh{} & \CoTRPNt{} & \ToTNL{} & \ToTNLh{} & \ToTCode{} & \ToTCodeh{} & \ToTRPN{} & \ToTRPNh{} & \CriticNL{} & \CriticCode{} & \CriticRPN{} & \OPROC{} & \OPRONL{} & \OPRO{} & \OPRORPN{} & \EvoRPNw{} \\
\midrule
\multirow{3}{*}{\rotatebox{90}{\textbf{r \& c}}} 
& VG & \textcolor{gray}{\textless 0} & \textcolor{gray}{0.89} & 1.61 & 1.29 & 1.46 & \textbf{1.95} & 1.61 & 0.95 & 1.58 & 1.18 & 1.77 & 1.22 & 1.30 & 1.80 & 1.47 & 1.32 & \cellcolor{blue3}\textbf{2.30} & 1.52 & \textbf{1.96} & 1.40 & \cellcolor{yellow3}0.64 & 2.02 & \cellcolor{blue1}\textbf{2.56} & \cellcolor{blue2}2.40 &  1.21\\
& TG & \textcolor{gray}{--} & \textcolor{gray}{0.85} & 1.79 & 1.40 & 1.55 & \textbf{1.92} & 1.67 & 1.06 & 1.59 & 1.19 & 1.86 & 1.40 & 1.27 & 1.84 & 1.39 & 1.42 & \cellcolor{blue3}\textbf{2.38} & 1.46 &\textbf{2.06} & 1.51 & \cellcolor{yellow3}0.26 & 1.87 & \cellcolor{blue1}\textbf{2.59} & \cellcolor{blue2}2.50 &  1.26\\
& AG & \textcolor{gray}{--} & \textcolor{gray}{-0.15} & \cellcolor{blue3}\textbf{0.31} & -1.02 & 0.21 & -1.26 & -1.73 & -0.30 & -0.27 & -0.22 & 0.08 & 0.15 & \cellcolor{blue2}0.32 & -0.01 & 0.20 & -0.13 & \cellcolor{blue1}\textbf{0.33} & 0.19 & \textbf{0.26} & 0.11 & -1.41 & \cellcolor{yellow3}-2.06 & -0.09 & \textbf{0.06} &  -0.93\\
\midrule
\multirow{3}{*}{\rotatebox{90}{\textbf{poker}}} 
& VG & \textcolor{gray}{--} & \textcolor{gray}{24.6} & 17.0 & 18.5 & 18.7 & \cellcolor{blue2}\textbf{26.0} & 25.6 & \cellcolor{blue2}\textbf{26.0} & 21.9 & 20.5 & 22.1 & 11.3 & 7.0 & \cellcolor{blue3}25.9 & \cellcolor{blue1}\textbf{26.1} & 22.6 & 16.9 & 9.7 & \textbf{25.0} & 17.9 & \cellcolor{yellow3}4.1 & 18.4 & \cellcolor{blue2}\textbf{26.0} & 24.1 &  24.8\\
& TG & \textcolor{gray}{--} & \textcolor{gray}{25.8} & 17.0 & 16.9 & 18.6 & \cellcolor{blue1}\textbf{26.1} & \cellcolor{blue3}25.6 & \cellcolor{blue1}\textbf{26.1} & 21.9 & 20.5 & 22.2 & 11.3 & 7.1 & \cellcolor{blue2}26.0 & \cellcolor{blue1}\textbf{26.1} & 22.7 & 16.8 & 9.7 &\textbf{25.0} & 18.0 & \cellcolor{yellow3}0.0 & 18.5 & \cellcolor{blue2}\textbf{26.0} & 24.2 &  24.9\\
& AG & \textcolor{gray}{--} & \textcolor{gray}{-0.2} & \cellcolor{yellow3}-16.8 & -0.1 & -1.1 & -2.3 & -0.5 & \cellcolor{blue2}0.0 & -4.4 & -0.2 & \cellcolor{blue1}\textbf{0.2} & -0.1 & \cellcolor{blue2}\textbf{0.0} & -0.1 & \cellcolor{blue2}\textbf{0.0} & -0.4 & -3.4 & \textbf{-0.1} & -1.1 & -0.2 & -0.3 & \textbf{-0.1} & \textbf{-0.1} & -1.1 &  -3.7\\
\bottomrule
\end{tabular}%
}
\end{table*}

\subsection{Scalability Analysis (\textbf{RQ 6})}\label{ssec:scale}
The datasets investigated in LATTE are generally relatively small, typically comprising around 10k instances. In this subsection, we analyze datasets ranging from 100k to 1M instances to verify whether our findings are influenced by data scale. The detailed results are presented in ~\Cref{tab:scale}.

\emph{\textbf{Finding 14.} \uline{Data scale naturally bridges the validation-test gap, neutralizing overfitting risks without algorithmic intervention.}}
As the instance count increases with feature and class dimensions held constant, the validation set captures the underlying distribution more effectively, allowing the validation model to compute scores with greater precision. Consequently, while data-scarce environments require rigorous algorithmic safeguards like Best-of-N selection (\emph{\textbf{Finding 3}}), sheer data volume inherently minimizes over-searching risks in LATTE.

\emph{\textbf{Finding 15.} \uline{Underlying task logic rigidly constrains optimal representation formats, overriding both data volume and prompt engineering.}}
By sampling fixed instances and relying on calculated metadata, LATTE remains insensitive to total data volume (e.g., covertype and road-safety trends mirror those of smaller datasets). Conversely, poker-hand requires complex logic (Texas Hold'em rules) for 100\% accuracy. Since NL and RPN formats struggle to encode such rules, the Code format supersedes prompting techniques and FE strategies, achieving optimal results across all families.

\begin{figure}
  \centering
  \includegraphics[width=0.8\linewidth]{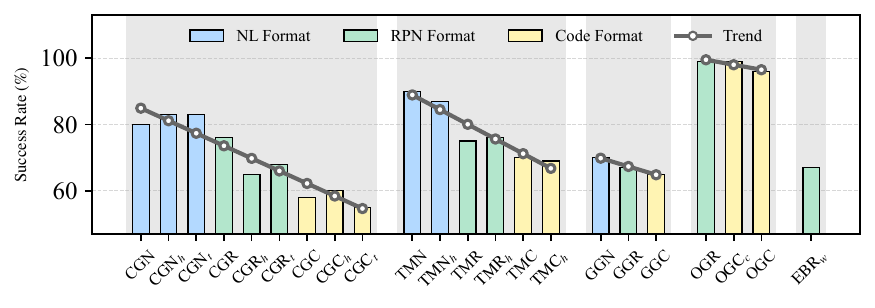}
  \caption{Success Rates across methods. \textbf{Reading Guide (\Cref{ssec:robust}):} Note the inverse relationship between format expressiveness and execution stability (\emph{Finding 16}), and the near-perfect robustness achieved by iterative OPRO optimization (\emph{Finding 17}).}
  \label{fig:rate}
\end{figure}

\subsection{Robustness Analysis (\textbf{RQ 7})}\label{ssec:robust}
This section examines the robustness of LATTE methods across datasets and random splits using success rate as the evaluation metric, as methods with higher success rates exhibit more stable performance. Results are presented in~\Cref{fig:rate}.

\emph{\textbf{Finding 16.} \uline{There is a direct trade-off between format expressiveness and execution stability: unrestricted structural flexibility triggers severe runtime hallucinations.}}
As illustrated in~\Cref{fig:rate}, a consistent trend emerges: RPN underperforms NL, while Code exhibits the lowest success rates. We observe that the flexibility of the Code format frequently induces problematic LLM behaviors, such as referencing non-existent operators or features, or even manipulating prediction targets to inflate performance metrics. These issues result in a substantial number of runtime errors.

\emph{\textbf{Finding 17.} \uline{Iterative optimization bottlenecks inherently regularize generation, achieving near-perfect execution stability.}}
\OPRORPN{}~and \OPROC{}~achieve the highest success rate of 99\%, attributed to their adoption of OPRO. In each iteration, these methods generate and refine just one new feature expression, thereby maintaining a focused and streamlined optimization process.

\section{Recommendations and Beyond}\label{sec:recommendation}

We introduce \LATTEArena{}, a comprehensive benchmark and evaluation framework designed to demystify LLM-powered Automated Tabular Feature Engineering. From our systematic evaluation of this sprawling design space, we distill three actionable, practitioner-focused recommendations for real-world deployment, organized by \emph{overall performance} (RQs 1-3), \emph{component design} (RQs 4-5), and \emph{scalability} (RQs 6-7):

\textbf{(I) Align search strategies with your budget, and output formats with your data and task.} Default to zero-shot RPN prompting under tight token budgets, tree-based exploration (ToT) for moderate budgets, and reserve iterative methods (OPRO with Best-of-N) for unlimited budgets. For formats, strictly use Code for regression (to handle complex math) and RPN for classification (for rapid, broad exploration).

\textbf{(II) Invest tokens in high-yield context, not convoluted pipelines.} Skip bulky demonstrations and raw calculated values. Instead, invest your context window in lightweight structural priors (e.g., CART). Crucially, \emph{never} bypass the downstream feature selector: it is your primary lever for balancing latency and accuracy. To cut costs further, aggressively trim metadata/samples before altering search logic.

\textbf{(III) Anchor robustness strategies to specific failure modes: overfitting, logic constraints, and runtime crashes.} On small datasets, counter overfitting by using Best-of-N sampling. If the task demands strict logic, enforce the Code format. Finally, to prevent runtime crashes from LLM hallucinations, rely on iterative refinement or rule-based error correction.

Our benchmarking identifies three critical bottlenecks: (1) no single method dominates across data scales; (2) algorithmic complexity yields diminishing returns; and (3) current demonstrations remain cost-ineffective.
Consequently, future improvements must pivot from intricate prompting toward optimizing tabular context management. Building on \LATTEArena{}, promising directions include advanced context retrieval, richer multi-dimensional scoring, and tabular-specific SFT/RL paradigms. We hope this work provides a solid foundation for advancing automated feature engineering.

\bibliographystyle{formatting/plainnat}
\bibliography{sample}

\end{document}